\documentclass[runningheads]{llncs}

\usepackage{macros}
\usepackage[dvipsnames]{xcolor}
\usepackage{svg}
\usepackage[T1]{fontenc}
\usepackage{graphicx}
\usepackage{wrapfig}
\usepackage{hyperref}
\usepackage[T1]{fontenc}
\begin{document}
%%%%%%%%%%%%%%%%%%%%%%%%%%%%%%%%%%%%%%%%%%%%%%%%%%%%%%%%%%%%%%%%%
%%%%%%%%%%%%%%%%%%%%%%%%%%%%%%%%%%%%%%%%%%%%%%%%%%%%%%%%%%%%%%%%%
\title{Small, Private Language Models as Teammates for Educational Assessment Design}
%%%%%%%%%%%%%%%%%%%%%%%%%%%%%%%%%%%%%%%%%%%%%%%%%%%%%%%%%%%%%%%%%
\titlerunning{Small, Private Language Models as Teammates}
%%%%%%%%%%%%%%%%%%%%%%%%%%%%%%%%%%%%%%%%%%%%%%%%%%%%%%%%%%%%%%%%%

\author{Chris Davis Jaldi\inst{1}\orcidID{0009-0000-2287-1198} \and
Anmol Saini\inst{1}\orcidID{0009-0000-1735-2377} \and
Shan Zhang\inst{2}\orcidID{0009-0003-3532-0661} \and
Noah Schroeder\inst{2}\orcidID{0000-0002-3281-2594} \and
Cogan Shimizu\inst{1}\orcidID{0000-0003-4283-8701} \and
Eleni Ilkou\inst{3}\orcidID{0000-0002-4847-6177}}

\authorrunning{C.D. Jaldi et al.}
%%%%%%%%%%%%%%%%%%%%%%%%%%%%%%%%%%%%%%%%%%%%%%%%%%%%%%%%%%%%%%%%
\institute{
Wright State University, USA\\
\email{\{jaldi.2, saini.25, cogan.shimizu\}@wright.edu} \and
University of Florida, USA\\
\email{\{zhangshan, schroedern\}@ufl.edu} \and
TIB – Leibniz Information Centre for Science and Technology, Germany\\
\email{eleni.ilkou@tib.eu}
}

\maketitle
%%%%%%%%%%%%%%%%%%%%%%%%%%%%%%%%%%%%%%%%%%%%%%%%%%%%%%%%%%%%%%%%%
\begin{abstract} 
Generative AI increasingly supports educational design tasks, e.g., through \emph{Large} Language Models (LLMs), demonstrating the capability to design assessment questions that are aligned with pedagogical frameworks (e.g., Bloom's taxonomy). However, they often rely on subjective or limited evaluation methods; focus primarily on proprietary models; or rarely systematically examine generation, evaluation, or deployment constraints in real educational settings. Meanwhile, \emph{Small} Language Models (SLMs) have emerged as local alternatives that better address privacy and resource limitations; yet their effectiveness for assessment tasks remains underexplored.
To address this gap, we systematically compare LLMs and SLMs for assessment question design; evaluate generation quality across Bloom's taxonomy levels using reproducible, pedagogically grounded metrics; and further assess model-based judging against expert-informed evaluation by analyzing reliability and agreement patterns.
Results show that SLMs achieve competitive performance across key pedagogically motivated quality dimensions while enabling local, privacy-sensitive deployment. However, model-based evaluations also exhibit systematic inconsistencies and bias relative to expert ratings. These findings provide evidence to posit language models as bounded assistants in assessment workflows; underscore the necessity of Human-in-the-Loop; and advance the automated educational question generation field by examining quality, reliability, and deployment-aware trade-offs.

\keywords{Small Language Models (SLMs) \and Automated Educational Question Generation (AEQG) \and Bloom's Taxonomy.}
\end{abstract}
%%%%%%%%%%%%%%%%%%%%%%%%%%%%%%%%%%%%%%%%%%%%%%%%%%%%%%%%%%%%%%%%%
%%%%%%%%%%%%%%%%%%%%%%%%%%%%%%%%%%%%%%%%%%%%%%%%%%%%%%%%%%%%%%%%%
\section{Introduction and Related Work}
\label{sec:intro}
%%%%%%%%%%%%%%%%%%%%%%%%
Recent advances in Generative AI (GenAI), particularly Large Language Models (LLMs), have led to an influx of LLMs being used across many educational settings, including tutoring~\cite{borchers2025can,stamper2024enhancing,vanzo2024gpt}, adaptive assessment~\cite{kabir2023llm,meyer2024using}, and other automated personalized learning systems~\cite{jaldi2025education}. Among these emerging uses, automated educational question generation (AEQG) has gained attention as one promising way to reduce instructor workload while supporting differentiated and formative assessment~\cite{bulathwela2023scalable,maity2024exploring}. Prior work demonstrates that LLMs can generate fluent questions aligned with pedagogical frameworks such as Bloom’s taxonomy~\cite{fawzi2024towards,hwang2024towards,wang2022towards}. However, evaluations often rely on subjective rubrics, limited statistical analysis, or proprietary cloud-based models, leaving open questions about reliability, reproducibility, and real-world deployability.

While research used template-based or sequence-to-sequence approaches, contemporary models substantially improved lexical coherence and variety. However, these highlight the challenge of controlling complexity or avoiding distractors ~\cite{elkins2023useful,raz2025automated}, reinforcing that even proprietary LLM results can drift from target learning outcomes without human post-intervention. Simultaneously, a growing line of work on Bloom's taxonomy-oriented AEQG shows that explicit cues can condition AI models to produce targeted questions~\cite{hwang2024towards,wang2022towards}. Few even demonstrated that pedagogically aligned prompting improved validity even when they still struggle with increasing complexity or semantic precision~\cite{fawzi2024towards,scaria2024automated}. AEQG evaluation remained a major open problem with most relying on human annotations~\cite{bulathwela2023scalable,wang2022towards} -- a costly and non-scalable approach -- or using AI as a judge, leading to other pitfalls~\cite{zheng2023judging}. In educational contexts, these settings raise further concerns about using AI for assessment design evaluation.

Beyond the capabilities of frontier LLMs, deployment realities further complicate their adoption. Many educational institutions face constraints around cost and students' data privacy, making the reliance on external APIs problematic. These concerns have motivated growing interest in Small Language Models (SLMs), light, often open-weight alternatives that can run locally and support privacy-preserving or resource-constrained environments~\cite{fawzi2024towards,schick2020s,vuruma2024cloud}. However, despite these advantages and early signs of deployment in educational settings~\cite{reza2025small}, their pedagogical capabilities remain underexplored for assessment design tasks that demand content validity, readability, and cognitive alignment~\cite{allison2025generative}; and systematic comparisons between proprietary models and local SLMs are scarce~\cite{ilkou2025dyslexia,reza2025small}. As a result, little is known about whether these models can provide comparable quality while offering safer, scalable deployment in under-resourced classroom workflows~\cite{zhang2024transforming}, where time and money are scarce. 

At the same time, the task itself also presents intrinsic challenges. Generating \textit{effective} assessment items requires more than linguistic fluency: Questions should align with learning objectives, reflect appropriate cognitive complexity, remain readable, and avoid misleading or extraneous content~\cite{elkins2023useful,raz2025automated}. Existing approaches often depend on costly human rubric-based ratings or automated model-based judgments whose reliability and potential biases are not well studied~\cite{bulathwela2023scalable,wang2022towards,zheng2023judging}. With that, both the quality of generated questions and the trustworthiness of automated evaluation remain open concerns. Thus, both generation quality and evaluation trust remain unresolved in the state-of-the-art (SOTA).

\begin{figure}[tb]
    \centering
    \includegraphics[width=0.85\textwidth]{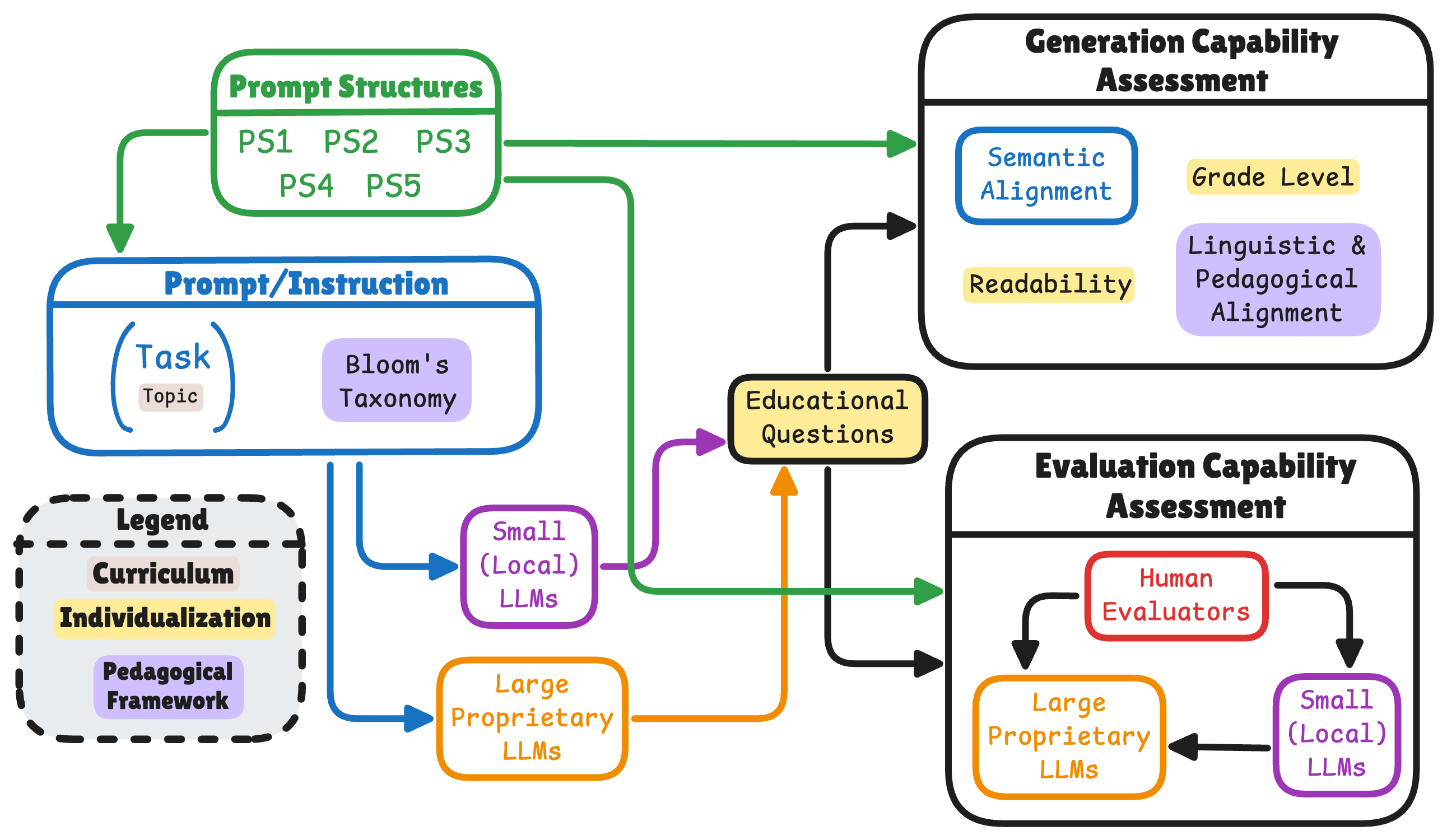}
    \vspace{-0.3cm}
    \caption{A graphical overview of the experimental workflow \& conceptual research roadmap.}
\label{figure:tech-exp}
\vspace{-0.55cm}
\end{figure}

To address these intertwined gaps -- generation quality, evaluation reliability, and deployment feasibility -- we built on Scaria et al.'s work~\cite{scaria2024automated}, which demonstrated Bloom-aligned question generation using human and model ratings with a rubric, and extended it with a more deployment-aware framework by systematically evaluating and comparing LLMs' and SLMs' performances using objective linguistic and semantic metrics. Drawing on established, reproducible Natural Language Processing (NLP) metrics with known correspondence to human judgments~\cite{raz2025automated,zhang2019bertscore} and conducting evaluation across multiple prompt structures (PS), we assessed quality with proxies for each against the core dimensions: \emph{cognitive complexity}, \emph{linguistic intent}, and \emph{pedagogical compliances}. Building on model-as-a-judge work~\cite{zheng2023judging}, we analyzed reliability, bias, consistency, and agreement patterns. Consistent with human-in-the-loop (HIL) perspectives~\cite{memarian2024human}, we posit language models (LMs) as bounded assistants rather than autonomous assessors. Fig~\ref{figure:tech-exp} outlines our conceptual experimental workflow and research roadmap to scope GenAI for controlled, auditable roles. We designed our study around four research questions (RQs), initially assessing generative-pedagogical compliances and then evaluative capabilities:

\small
\begin{compactenum} [\bf RQ1.]
    \item To what extent do LMs maintain targeted and appropriate complexity (grade level and readability across prompts \& Bloom's taxonomy levels)?
    \item To what extent do generated questions maintain intent-aligned topical relevance and consistency as they vary across Bloom’s levels within topics?
    \item How well do LMs align the generated question to each targeted taxonomy level (consistent action verb usage)?
    \item How well do LMs assess the quality of generated questions, and is this self-evaluation reliable?
\end{compactenum}
\normalsize

%%%%%%%%%%%%%%%%%%%%%%%%%%%%%%%%%%%%%%%%%%
%%%%%%%%%%%%%%%%%%%%%%%%%%%%%%%%%%%%%%%%%%

Building on but extending beyond Scaria et al.'s AEQG work~\cite{scaria2024automated}, our study, in contrast, contributes by:
\begin{inparaenum}[\bf (a)]
    \item systematically benchmarking SLMs alongside LLMs;
    \item introducing reproducible objective metrics across linguistic, semantic, and pedagogically grounded quality dimensions;
    \item solidifying insights with statistical rigor; and
    \item thoroughly analyzing model-as-a-judge reliability and bias patterns.
\end{inparaenum}
These comparative analyses close key gaps at the intersection of AEQG and LMs, assessing their strengths and limitations and laying a foundation for future classroom-grounded, curriculum-aware GenAI systems.

%%%%%%%%%%%%%%%%%%%%%%%%%%%%%%%%%%%%%%%%%%%%%%%%%%%%%%%%%%%%%%%%%
\section{Methodology}
\label{sec:meth}
%%%%%%%%%%%%%%%%%%%%%%%%%%%%%%%%%%%%%%%%%%%%%%%%%%%%%%%%%%%%%%%%%
We reused inputs (2550 questions) from~\cite{scaria2024automated} to preserve comparability, additionally generating 4080 new ones to isolate model constraint effects, systematically evaluate capabilities and trustworthiness, and use inferential statistics.

%%%%%%%%%%%%%%%%%%%%%%%%%%%%%%%%%%%%%%%%%%%%%%%%%%%%%%%%%%%%%%%%%
\subsection{Study Design}
%%%%%%%%%%%%%%%%%%%%%%%%%%%%%%%%%%%%%%%%%%%%%%%%%%%%%%%%%%%%%%%%%
Our end-to-end workflow design is summarized in Fig~\ref{figure:tech-exp}. We have further introduced compliance checks for targeting grade-level inference, action verb usage, and cohesion between questions to better align with pedagogical expectations, bridging technical NLP evaluation with education research. We assess two model families: frontier models and locally deployable alternatives.

\noindent\textbf{Models Used.} For LLMs, we reused the exact questions from the original study's dataset. Models include:
\begin{inparaenum} [\it i.]
    \item GPT-4 (G4),
    \item GPT-3.5 (G3.5),
    \item Palm-2 (P2)
    \item LLaMA-2 (L2),
    \item Mistral (M), and
    \item Gemini Pro (GP) -- as evaluator.
\end{inparaenum}
For SLMs, we prioritized open-weight, local, and low-resource deployable alternatives for resource-scarce environments such as schools, but still deliver SOTA performance. They are:
\begin{inparaenum} [\it i.]
    \item Granite 4 (Gr4),
    \item Phi-4 mini (P4m),
    \item LLaMA 3.2 (L3.2),
    \item Mistral Small 3.2 (MS),
    \item GPT-OSS (GO),
    \item Phi-4 (P4),
    \item Deepseek R1 (D1), and
    \item Gemma 3 (Ge3).
\end{inparaenum}
Generation and evaluation temperatures are fixed at \emph{0.8} and \emph{0.0} (consistent with the original study), enabling direct comparison between proprietary LLMs and open-source SLMs. Additional configuration details are available in \cite{site:git-repo}.

%%%%%%%%%%%%%%%%%%%%%%%%%%%%%%%%%%%%%%%%%%%
\noindent\textbf{Topics and Prompt Strategy.} We adopted the prompt framework and topics from Scaria et al.~\cite{scaria2024automated} to systematically elicit questions across Bloom's taxonomy levels while ensuring consistent comparability. This encompassed 17 Machine Learning and Data Science topics. Five PS were used: \textbf{PS1:} general question generation with no educational specifications (zero-shot prompting) and \textbf{PS2--5:} incrementally more specific prompts targeting a specific educational level (graduate student) and demographic group (Indian) in question generation. These consisted of several components, such as \emph{system instruction}, \emph{few-shot or extended explicit cues}, \emph{task}, or \emph{additional targets}.

%%%%%%%%%%%%%%%%%%%%%%%%%%%%%%%%%%%%%%%%%%%
\subsection{Evaluation Metrics}
%%%%%
To conduct a thorough, systematic evaluation of the responses and not completely rely on subjective ratings, we paired each RQ with objective, reproducible NLP or compliance metrics to evaluate LMs' quality, complemented with statistical testing. We further performed multi-factor analysis across PS and models to check for model behavior, using inferential tests described later.

\noindent{\textbf{Cognitive Complexity Compliance} \emph{(with Grade Level and Reading Ease as proxies)} - RQ1.} We measured the grade level (U.S grade scale) and readability of generated questions with Flesch-Kincaid Grade Level~\cite{flesch1950measuring} and Flesch Reading Ease score~\cite{flesch1948new}, respectively, to assess text complexity. A low grade level and high reading score map to simpler text and vice versa. We used these metrics to
\begin{inparaenum}[\bf(a)]
    \item characterize the default complexity the models target,
    \item track how readability shifts across Bloom levels, and
    \item measure their responsiveness to prompts with explicit increased specificity.
\end{inparaenum}

\noindent{\textbf{Linguistic Intent Compliance} \emph{(with Prompt-topic and Within-topic Consistency as proxies)} - RQ2.} We measured topical and task intent adherence by computing semantic similarity between the input prompt and each generated question and within-topic consistency across six levels using BERTScore, which compared candidate and reference texts using contextual token embeddings~\cite{zhang2019bertscore}, to ensure responses did not waver. Aligning with SOTA methods for NLP tasks across domains~\cite{jaldi2025impact}, we leverage sentence transformer-based embeddings (GTR T5-large) to compute intent scores. Additionally, we computed pairwise scores across Bloom levels to capture whether models preserved coherent linguistic and semantic style while shifting cognitive demands for a more focused internal linguistic alignment check~\cite{zhang2024semi}.

\noindent{\textbf{Pedagogical Compliance} \emph{(with Verb Check as a proxy)} - RQ3.} We assessed compliance with Bloom's taxonomy by checking whether each question used an action verb consistent with its level~\cite{adams2015bloom}. We curated a verb list and scored each question as compliant or not. We aggregated compliance proportions by model and PS. This metric allowed us to capture whether models followed pedagogical conventions or drifted lexically from the instructional design framework.

\noindent{\textbf{Trustworthiness of Model Judgments} \emph{(with Agreement Check as a proxy) - RQ4.}
For judgment and trustworthiness assessment, we adapted the same rubric evaluation strategy as in~\cite{scaria2024automated} by extending it in two critical ways:
\begin{inparaenum}
    \item use of SLMs as evaluators on LLM-generated questions and
    \item comparison of evaluator outputs across multiple models against human expert annotations for reliability and bias patterns.
\end{inparaenum}
A question was categorized as a \emph{Quality} question if it was marked:
\begin{inparaenum}
    \item ``yes'' for \textit{Understandable}, \textit{Grammatical}, \textit{Clear}, and \textit{Answerable} and ``yes'' or ``maybe'' for \textit{WouldYouUseIt} or
    \item ``yes'' for \textit{Understandable}, \textit{Grammatical}, \textit{Rephrase}, and \textit{Answerable} and ``yes'' or ``more\_or\_less'' for \textit{Clear}.
\end{inparaenum}

We combined descriptive statistics and inferential testing, including means ($\mu$) to capture stability/drifts; assessed binary coded compliance; and tested differences across models and PS with ANOVA, formally evaluating group equality via F-test. All analyses, supplementary materials, prompts, generated question datasets and their statistics, additional details, and helper scripts are in a publicly accessible repository~\cite{site:git-repo} to facilitate replication and further research.

%%%%%%%%%%%%%%%%%%%%%%%%%%%%%%%%%%%%%%%%%%%%%%%%%%%%%%%%%%%%%%%%%
\section{Results}
\label{sec:results}
%%%%%
We present trends, statistical effects, and report empirical results organized by RQ for direct interpretation against the introduced evaluation framework.

%%%%%
\begin{table}[t]
\tiny
\centering
\setlength{\tabcolsep}{1pt}
\caption{Aggregated RQ results. 
\emph{$A_d$}=Adherence
{\tiny(\%)},
\emph{$G_L$}=Grade level,
\emph{$G_A$}=Grade Alignment
{\tiny(\%)},
\emph{$R_E$}=Reading Ease,
\emph{$B_A$}=BERT Alignment
{\tiny(\%)},
\emph{$V_A$}=Verb Alignment
{\tiny(\%)}
}
\label{table:all-results-consol}
\vspace{-0.2cm}
\begin{tabular}{l|clcccccccccccccccccccccc}
\multirow{14}{*}{\textbf{L}} & \multicolumn{1}{c|}{\multirow{2}{*}{\textbf{Model}}} & \multicolumn{5}{c|}{\textbf{PS1}} & \multicolumn{6}{c|}{\textbf{PS2}} & \multicolumn{6}{c|}{\textbf{PS3}} & \multicolumn{6}{c}{\textbf{PS4}} \\ \cline{3-25} 
 & \multicolumn{1}{c|}{} & \multicolumn{1}{c}{\textbf{$A_d$}} & \textbf{$G_L$} & \textbf{$R_E$} & \textbf{$B_A$} & \multicolumn{1}{c|}{\textbf{$V_A$}} & \textbf{$A_d$} & \textbf{$G_L$} & \textbf{$G_A$} & \textbf{$R_E$} & \textbf{$B_A$} & \multicolumn{1}{c|}{\textbf{$V_A$}} & \textbf{$A_d$} & \textbf{$G_L$} & \textbf{$G_A$} & \textbf{$R_E$} & \textbf{$B_A$} & \multicolumn{1}{c|}{\textbf{$V_A$}} & \textbf{$A_d$} & \textbf{$G_L$} & \textbf{$G_A$} & \textbf{$R_E$} & \textbf{$B_A$} & \textbf{$V_A$} \\ [1pt] \cline{2-25} 
 & \multicolumn{1}{c|}{G4} & \multicolumn{1}{c}{\textbf{65}} & 12 & 37 & \textbf{82} & \multicolumn{1}{c|}{76} & 65 & 16 & \textbf{68} & 20 & \textbf{88} & \multicolumn{1}{c|}{\textbf{98}} & 68 & \textbf{14} & 36 & 28 & 82 & \multicolumn{1}{c|}{\textbf{100}} & 65 & 15 & 50 & 26 & 82 & \textbf{100} \\
 & \multicolumn{1}{c|}{G3.5} & \multicolumn{1}{c}{\textbf{65}} & 15 & 21 & 59 & \multicolumn{1}{c|}{\textbf{79}} & \textbf{66} & 15 & 63 & 20 & 76 & \multicolumn{1}{c|}{93} & 68 & 16 & \textbf{65} & 20 & \textbf{88} & \multicolumn{1}{c|}{91} & \textbf{66} & 16 & \textbf{65} & 18 & \textbf{88} & \textbf{100} \\
 & \multicolumn{1}{c|}{P2} & \multicolumn{1}{c}{64} & \textbf{11} & \textbf{42} & 71 & \multicolumn{1}{c|}{62} & 65 & \textbf{13} & 34 & 30 & 53 & \multicolumn{1}{c|}{71} & 67 & \textbf{14} & 49 & 23 & 65 & \multicolumn{1}{c|}{75} & 64 & \textbf{14} & 45 & 28 & 59 & 80 \\
 & \multicolumn{1}{c|}{L2} & \multicolumn{1}{c}{\textbf{65}} & 12 & 37 & 41 & \multicolumn{1}{c|}{62} & 64 & \textbf{13} & 31 & \textbf{33} & 41 & \multicolumn{1}{c|}{91} & 67 & \textbf{14} & 41 & \textbf{34} & 47 & \multicolumn{1}{c|}{82} & 64 & \textbf{14} & 33 & \textbf{30} & 53 & 96 \\
 & \multicolumn{1}{c|}{M} & \multicolumn{1}{c}{\textbf{65}} & 13 & 29 & 71 & \multicolumn{1}{c|}{54} & 65 & 14 & 45 & 28 & 59 & \multicolumn{1}{c|}{75} & \textbf{69} & 15 & 55 & 22 & 53 & \multicolumn{1}{c|}{73} & 65 & 16 & 60 & 25 & 82 & 94 \\ \cline{4-22}
 & \multicolumn{1}{l}{} &  & \multicolumn{1}{l|}{\multirow{2}{*}{\textbf{Model}}} & \multicolumn{6}{c|}{\textbf{PS5}} & \multicolumn{6}{c|}{\textbf{$\mu$}} & \multicolumn{6}{c}{\textbf{CV}} &  &  &  \\ \cline{5-22}
 &  &  & \multicolumn{1}{l|}{} & \textbf{$A_d$} & \textbf{$G_L$} & \textbf{$G_A$} & \textbf{$R_E$} & \textbf{$B_A$} & \multicolumn{1}{c|}{\textbf{$V_A$}} & \textbf{$A_d$} & \textbf{$G_L$} & \textbf{$G_A$} & \textbf{$R_E$} & \textbf{$B_A$} & \multicolumn{1}{c|}{\textbf{$V_A$}} & \textbf{$A_d$} & \textbf{$G_L$} & \textbf{$G_A$} & \textbf{$R_E$} & \textbf{$B_A$} & \textbf{$V_A$} & \textbf{} & \multicolumn{1}{l}{} & \multicolumn{1}{l}{} \\ [1pt] \cline{4-22}
 &  &  & \multicolumn{1}{c|}{G4} & \textbf{66} & \textbf{14} & 37 & 29 & \textbf{94} & \multicolumn{1}{c|}{\textbf{100}} & \textbf{66} & 14 & 48 & 28 & \textbf{86} & \multicolumn{1}{c|}{\textbf{95}} & 2 & 10 & 27 & 19 & 5 & 10 &  & \multicolumn{1}{l}{} & \multicolumn{1}{l}{} \\
 &  &  & \multicolumn{1}{c|}{G3.5} & 65 & 16 & 63 & 20 & 76 & \multicolumn{1}{c|}{\textbf{100}} & \textbf{66} & 16 & \textbf{64} & 20 & 78 & \multicolumn{1}{c|}{93} & 2 & 3 & 2 & 5 & 14 & 8 &  & \multicolumn{1}{l}{} & \multicolumn{1}{l}{} \\
 &  &  & \multicolumn{1}{c|}{P2} & 65 & 17 & \textbf{73} & 10 & 59 & \multicolumn{1}{c|}{91} & 65 & 14 & 50 & 27 & 61 & \multicolumn{1}{c|}{76} & 1 & 14 & 28 & 38 & 10 & 13 &  & \multicolumn{1}{l}{} & \multicolumn{1}{l}{} \\
 &  &  & \multicolumn{1}{c|}{L2} & 65 & \textbf{14} & 42 & \textbf{33} & 41 & \multicolumn{1}{c|}{97} & 65 & \textbf{13} & 37 & \textbf{33} & 45 & \multicolumn{1}{c|}{86} & 2 & 6 & 13 & 7 & 11 & 15 &  & \multicolumn{1}{l}{} & \multicolumn{1}{l}{} \\
 &  &  & \multicolumn{1}{c|}{M} & 65 & 15 & 54 & 26 & 76 & \multicolumn{1}{c|}{92} & \textbf{66} & 14 & 53 & 26 & 68 & \multicolumn{1}{c|}{78} & 2 & 6 & 10 & 10 & 16 & 19 &  & \multicolumn{1}{l}{} & \multicolumn{1}{l}{} \\ \hline
\multirow{20}{*}{\textbf{S}} & \multicolumn{1}{c|}{\multirow{2}{*}{\textbf{Model}}} & \multicolumn{5}{c|}{\textbf{PS1}} & \multicolumn{6}{c|}{\textbf{PS2}} & \multicolumn{6}{c|}{\textbf{PS3}} & \multicolumn{6}{c}{\textbf{PS4}} \\ \cline{3-25} 
 & \multicolumn{1}{c|}{} & \multicolumn{1}{c}{\textbf{$A_d$}} & \textbf{$G_L$} & \textbf{$R_E$} & \textbf{$B_A$} & \multicolumn{1}{c|}{\textbf{$V_A$}} & \textbf{$A_d$} & \textbf{$G_L$} & \textbf{$G_A$} & \textbf{$R_E$} & \textbf{$B_A$} & \multicolumn{1}{c|}{\textbf{$V_A$}} & \textbf{$A_d$} & \textbf{$G_L$} & \textbf{$G_A$} & \textbf{$R_E$} & \textbf{$B_A$} & \multicolumn{1}{c|}{\textbf{$V_A$}} & \textbf{$A_d$} & \textbf{$G_L$} & \textbf{$G_A$} & \textbf{$R_E$} & \textbf{$B_A$} & \textbf{$V_A$} \\ [1pt] \cline{2-25} 
 & \multicolumn{1}{c|}{Gr4} & \multicolumn{1}{c}{\textbf{69}} & 15 & 27 & 76 & \multicolumn{1}{c|}{76} & 65 & 14 & 48 & 30 & 88 & \multicolumn{1}{c|}{85} & 76 & 14 & 49 & 26 & 65 & \multicolumn{1}{c|}{75} & \textbf{79} & 15 & 51 & 22 & 47 & 88 \\
 & \multicolumn{1}{c|}{P4m} & \multicolumn{1}{c}{68} & 16 & 21 & \textbf{100} & \multicolumn{1}{c|}{82} & 65 & 16 & 62 & 20 & 82 & \multicolumn{1}{c|}{88} & 74 & 18 & 66 & 12 & 88 & \multicolumn{1}{c|}{81} & 77 & 16 & 56 & 19 & 80 & 78 \\
 & \multicolumn{1}{c|}{L3.2} & \multicolumn{1}{c}{\textbf{69}} & \textbf{12} & 37 & 71 & \multicolumn{1}{c|}{83} & 67 & 14 & 36 & 28 & 76 & \multicolumn{1}{c|}{86} & 76 & 16 & 64 & 19 & \textbf{94} & \multicolumn{1}{c|}{89} & 78 & 15 & 60 & 19 & 88 & 92 \\
 & \multicolumn{1}{c|}{MS} & \multicolumn{1}{c}{68} & \textbf{12} & \textbf{41} & 82 & \multicolumn{1}{c|}{80} & 66 & \textbf{13} & 32 & \textbf{32} & 71 & \multicolumn{1}{c|}{82} & \textbf{77} & 15 & 51 & 26 & 82 & \multicolumn{1}{c|}{84} & 76 & \textbf{14} & 38 & \textbf{30} & 76 & 91 \\
 & \multicolumn{1}{c|}{GO} & \multicolumn{1}{c}{66} & 13 & 37 & 71 & \multicolumn{1}{c|}{86} & 65 & \textbf{13} & 35 & \textbf{32} & 76 & \multicolumn{1}{c|}{88} & 72 & 14 & 43 & 27 & 88 & \multicolumn{1}{c|}{93} & 72 & 15 & 49 & 23 & 94 & 94 \\
 & \multicolumn{1}{c|}{P4} & \multicolumn{1}{c}{\textbf{69}} & 14 & 24 & 88 & \multicolumn{1}{c|}{87} & 67 & 16 & \textbf{69} & 17 & \textbf{100} & \multicolumn{1}{c|}{89} & \textbf{77} & 16 & \textbf{70} & 17 & 88 & \multicolumn{1}{c|}{92} & 76 & 16 & \textbf{67} & 18 & 82 & 86 \\
 & \multicolumn{1}{c|}{D1} & \multicolumn{1}{c}{67} & 13 & 31 & 71 & \multicolumn{1}{c|}{72} & 66 & \textbf{13} & 33 & 27 & 76 & \multicolumn{1}{c|}{89} & 75 & 14 & 42 & 25 & 53 & \multicolumn{1}{c|}{85} & 76 & \textbf{14} & 44 & 25 & 76 & 82 \\
 & \multicolumn{1}{c|}{Ge3} & \multicolumn{1}{c}{\textbf{69}} & 13 & 37 & \textbf{100} & \multicolumn{1}{c|}{\textbf{97}} & \textbf{68} & 14 & 29 & 28 & \textbf{100} & \multicolumn{1}{c|}{\textbf{96}} & 74 & \textbf{13} & 30 & \textbf{29} & 88 & \multicolumn{1}{c|}{\textbf{95}} & 75 & \textbf{14} & 41 & 25 & \textbf{100} & \textbf{96} \\ \cline{4-22}
 & \multicolumn{1}{l}{} &  & \multicolumn{1}{l|}{\multirow{2}{*}{\textbf{Model}}} & \multicolumn{6}{c|}{\textbf{PS5}} & \multicolumn{6}{c|}{\textbf{$\mu$}} & \multicolumn{6}{c}{\textbf{CV}} &  &  &  \\ \cline{5-22}
 &  &  & \multicolumn{1}{l|}{} & \textbf{$A_d$} & \textbf{$G_L$} & \textbf{$G_A$} & \textbf{$R_E$} & \textbf{$B_A$} & \multicolumn{1}{c|}{\textbf{$V_A$}} & \textbf{$A_d$} & \textbf{$G_L$} & \textbf{$G_A$} & \textbf{$R_E$} & \textbf{$B_A$} & \multicolumn{1}{c|}{\textbf{$V_A$}} & \textbf{$A_d$} & \textbf{$G_L$} & \textbf{$G_A$} & \textbf{$R_E$} & \textbf{$B_A$} & \textbf{$V_A$} & \textbf{} & \multicolumn{1}{l}{} & \multicolumn{1}{l}{} \\ [1pt] \cline{4-22}
 &  &  & \multicolumn{1}{c|}{Gr4} & 77 & \textbf{14} & 41 & \textbf{28} & 65 & \multicolumn{1}{c|}{84} & \textbf{73} & \textbf{14} & 47 & 27 & 68 & \multicolumn{1}{c|}{82} & 7 & 2 & 8 & 10 & 20 & 7 &  & \multicolumn{1}{l}{} & \multicolumn{1}{l}{} \\
 &  &  & \multicolumn{1}{c|}{P4m} & 76 & 16 & 57 & 22 & 76 & \multicolumn{1}{c|}{80} & 72 & 16 & 60 & 19 & 86 & \multicolumn{1}{c|}{82} & 6 & 4 & 6 & 18 & 10 & 4 &  & \multicolumn{1}{l}{} & \multicolumn{1}{l}{} \\
 &  &  & \multicolumn{1}{c|}{L3.2} & 77 & \textbf{14} & 44 & 24 & \textbf{94} & \multicolumn{1}{c|}{93} & \textbf{73} & \textbf{14} & 51 & 25 & 85 & \multicolumn{1}{c|}{89} & 6 & 9 & 22 & 27 & 11 & 4 &  & \multicolumn{1}{l}{} & \multicolumn{1}{l}{} \\
 &  &  & \multicolumn{1}{c|}{MS} & 77 & \textbf{14} & 38 & \textbf{28} & \textbf{94} & \multicolumn{1}{c|}{90} & \textbf{73} & \textbf{14} & 40 & \textbf{31} & 81 & \multicolumn{1}{c|}{86} & 7 & 7 & 17 & 17 & 10 & 5 &  & \multicolumn{1}{l}{} & \multicolumn{1}{l}{} \\
 &  &  & \multicolumn{1}{c|}{GO} & 73 & \textbf{14} & 46 & 23 & \textbf{94} & \multicolumn{1}{c|}{94} & 70 & \textbf{14} & 43 & 28 & 85 & \multicolumn{1}{c|}{91} & 5 & 5 & 12 & 19 & 11 & 4 &  & \multicolumn{1}{l}{} & \multicolumn{1}{l}{} \\
 &  &  & \multicolumn{1}{c|}{P4} & 77 & 16 & \textbf{71} & 17 & \textbf{94} & \multicolumn{1}{c|}{89} & \textbf{73} & 16 & \textbf{69} & 19 & 91 & \multicolumn{1}{c|}{91} & 6 & 4 & 2 & 16 & \textbf{7} & 5 &  & \multicolumn{1}{l}{} & \multicolumn{1}{l}{} \\
 &  &  & \multicolumn{1}{c|}{D1} & 76 & \textbf{14} & 41 & \textbf{28} & 71 & \multicolumn{1}{c|}{85} & 72 & \textbf{14} & 40 & 27 & 69 & \multicolumn{1}{c|}{83} & 6 & 3 & 10 & 8 & 12 & 7 &  & \multicolumn{1}{l}{} & \multicolumn{1}{l}{} \\
 &  &  & \multicolumn{1}{c|}{Ge3} & \textbf{78} & \textbf{14} & 50 & 24 & 88 & \multicolumn{1}{c|}{\textbf{98}} & \textbf{73} & \textbf{14} & 38 & 29 & \textbf{95} & \multicolumn{1}{c|}{\textbf{96}} & 5 & 5 & 22 & 16 & 6 & 1 &  & \multicolumn{1}{l}{} & \multicolumn{1}{l}{}
\end{tabular}
\vspace{-0.5cm}
\end{table}
%%%%%%%%%%%%%%%%%%%%%%%%

\noindent{\textbf{RQ1 (Cognitive Complexity Compliance)}.} We assessed the models' consistency through \emph{default grade level ($G_L$)} of model outputs, \emph{grade-level compliance ($G_A$)} under targeted prompts, and \emph{readability ($R_E$)} (Table~\ref{table:all-results-consol}).

\noindent\emph{Grade Level.} Both LLMs and SLMs produced questions that corresponded to late high-school or early undergraduate levels ($\mu\approx14$). Overall, models performed comparably, with ANOVA showing no significant differences in default grade level ($F(1,11)=0.3$, $p=.6$). However, within-families revealed variations with both model and PS being significantly affected for SLMs ($F_{\text{models}}(7,28)=10.4$, $F_{\text{PS}}(4,28)=5.6$, $p<.002$) and LLMs ($F_{\text{models}}(4,16)=3.6$, $F_{\text{PS}}(4,16)=5.4$, $p<.03$). Overall, with greater uniformity across models, this indicates that both families aggregately converged to similar baseline grade complexity; but the architectural differences and instructional cues jointly influenced grade complexity while also indicating a similar sensitivity to prompt design. We further assessed compliance check by question across PS that explicitly specified a target level (PS2--5). Both families showed increased alignment with grade levels across prompts. However, patterns differed with only model having significant effect ($F_{\text{models}}(7,21)=11.2$, $p<.001$, $F_{\text{PS}}(3,21)=2.8$, $p=.07$) for SLMs and none for LLMs ($F_{\text{models}}(4,12)=2.9$, $F_{\text{PS}}(3,12)=0.2$, $p>.07$), reflecting a more homogeneous but less adaptive behavior.

\noindent\emph{Reading Ease.} We next examined readability and whether it fluctuates across levels. LLMs and SLMs exhibited comparable ($\mu\approx26$) aggregate scores with no differences ($F(1,11)=0.2$, $p=.7$). However, within-family analyses revealed distinct patterns, with SLMs demonstrating significant effects of both model and PS ($F_{\text{models}}(7,28)=10.1$, $F_{\text{PS}}(4,28)=12.0$, $p<.001$) whereas LLMs differed significantly across models ($F(4,16)=4.3$, $p=.02$) but not PS ($F(4,16)=2.4$, $p=.10$). Readability generally decreased as PS became more cognitively demanding, especially for Bloom levels \textit{analyze}, \textit{evaluate}, and \textit{create}. However, the magnitude and consistency differed: SLMs showed clearer and more systematic modulation whereas LLMs exhibited greater variability and less consistent adaptation. This trend confirmed the anticipated behavior that as the pedagogy target (Bloom's level) rose, the language became harder to read.

These findings suggest that while both families generate questions with comparable readability and grade levels, SLMs demonstrated greater control in adjusting language complexity in line with the intended pedagogical technique and grade instructional cues. This is especially relevant for educational settings where instructors must balance cognitive challenge with linguistic accessibility.

%%%%%%%%%%%%%%%%%%%%%%%%%%%
\begin{figure}[pt]
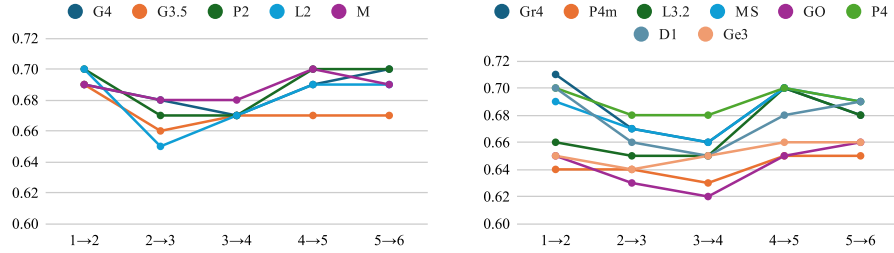

    \centering
    \begin{minipage}[t]{0.48\textwidth}
        \centering
        \includesvg[width=\linewidth]{figures/MEAN_LLM_BERT.svg}
    \end{minipage}\hfill
    \begin{minipage}[t]{0.48\textwidth}
        \centering
        \includesvg[width=\linewidth]{figures/MEAN_SLM_BERT.svg}
    \end{minipage}
    \vspace{-0.3cm}
    \caption{Bloom's level transition trend with LLMs on the left and SLMs on the right; x-axis represents level transitions and y-axis the avg BERTScore F1.}
    \label{figure:bloom-transition-trend}
    \vspace{-0.5cm}
\end{figure}
%%%%%%%%%%%%%%%%%%%%%%%%%%%

\noindent{\textbf{RQ2 (Linguistic Intent Compliance)}.} 
We evaluated the degree to which the questions generated by LLMs and SLMs remain topically aligned with respect to the input task query ($A_d$) (Table~\ref{table:all-results-consol}). Across all PS, both families produced questions that remained strongly aligned with the input topics, also showing a trend with increasing specificity and cognitive complexity. ANOVA comparing them showed a significant difference ($F(1,11)=171.2$, $p<.001$), indicating that SLMs maintained closer topical intent alignment. Further investigation of variability with model family and PS revealed significant effects for SLMs ($F_{\text{models}}(7,28)=6.8$, $F_{\text{PS}}(4,28)=167.9$, $p<.001$) and LLMs ($F_{\text{models}}(4,16)=4.9$, $F_{\text{PS}}(4,16)=31.8$, $p<.009$). In both families, the more structured PS (PS3--5), which explicitly encoded Bloom levels and additional constraints, yielded higher adherence than the simpler topic-only prompts (PS1--2). We continued to evaluate the linguistic consistency across successive levels ($1\xrightarrow~2,\ 2\xrightarrow~3$, and so on) using BERTScore (precision, recall, and F1) computed pairwise within the same topic, to capture complementary aspects of semantic stability as cognitive demand increases. Both model families maintained high F1 scores at early transition ($1\xrightarrow~2$) and distinct patterns at mid-levels but with identical aggregations overall (see Fig~\ref{figure:bloom-transition-trend}). However, SLMs exhibited flatter trajectories, indicating steadier linguistic progression. LLMs' precision score ($\mu=54.6$) revealed a more conceptual drift with Bloom's level than SLMs ($\mu=69.8$). Similarly, recall of LLMs ($\mu=56.7$) reflected increased risk of semantic omission over SLMs ($\mu=68.4$) and F1 the overall semantic stability of SLMs ($\mu=82.5$) over LLMs ($\mu=67.5$). Though significance is only observed for F1, SLMs had performance comparable to or sometimes greater than LLMs. Variance analysis revealed important differences where LLMs exhibited higher variability across transitions while SLMs maintained lower CVs. Precision, recall and F1 scores consistently exhibited significant effect of model (Small: $F_p(7,28)=6.5, F_r(7,28)=5.7, F_\text{F1}(7,28)=4.00, p<.004$; Large: $F_p(4,16)=11.7, F_r(4,16)=8.2, F_\text{F1}(4,16)=15.5, p<.0009$) but not PS, indicating that effects stemmed more from model behavior.

These results showed that 
\begin{inparaenum} [(i)]
    \item all LMs generally preserved the intent;
    \item SLMs achieved higher semantic adherence scores and were less prone to conceptual drift across Bloom's levels;
    \item while both families achieve high alignment across Bloom levels, the nature of their errors differed; and
    \item carefully designed PS substantially improved alignment for both families.
\end{inparaenum}
These illustrate how an instructor articulating generation instructions affects the overall question stability.

\noindent{\textbf{RQ3 (Pedagogical Compliance)}.} We evaluated whether all questions were aligned with their expected Bloom's level, with a verb compliance percentage score ($V_A$) per model and PS (Table~\ref{table:all-results-consol}). This captured the lexical markers of the pedagogical technique and instructions beyond surface-level fluency. Both model families demonstrated strong compliance, indicating that LMs can internalize and operationalize explicit pedagogical constraints. However, important structural differences emerged:
LLMs' verb compliance increased with increased PS specificity, with high compliance in all PS, though comparatively lower in earlier ones (PS1--3), where Bloom-level cues were not explicitly reinforced. ANOVA further confirmed significant effects of both model and PS ($F_{\text{models}}(4,16)=14.6$, $F_{\text{PS}}(4,16)=26.5$, $p<.001$). This pattern indicates that LLMs remained strongly effective at maintaining pedagogical alignment with explicit instructional scaffolding but less stable under minimal underspecified prompts. Such sensitivity raises concerns for autonomous deployment, where prompt formulation may propagate pedagogical misalignment. On the contrary, SLMs exhibited consistently high verb alignment even across lower-specificity prompts, underscoring the potential for pedagogical applications. When statistically analyzed, a significant effect was observed by model ($F=8.3, p<.001$), but not as significant by PS ($F=3.2, p=.03$). This further indicated that SLMs maintained consistency with pedagogical constraints more uniformly regardless of varied instructional conditions (prompt intricacies), whereas LLMs demonstrate strong pedagogical compliance only when heavily scaffolded. However, neither consistently guarantees perfect pedagogical correctness in isolation.

Cross-metric correlation analysis revealed Verb Alignment's strong positive correlation with grade measures, suggesting that Bloom’s verb usage is tightly coupled with difficulty escalation in LLMs. In contrast, the same pattern was weaker in SLMs, indicating that SLMs often maintain verb-level compliance without proportionally inflating complexity. Additionally, BERT alignment metrics were weakly correlated with difficulty measures, despite exhibiting high internal coherence. These decoupling correlation patterns suggest that models can preserve semantic intent while varying cognitive complexity, reinforcing the need for human oversight in automated evaluations and highlighting systematic differences in how model families operationalize pedagogical constraints, with implications for reliability, transparency, and deployment in educational settings.

\noindent\textbf{RQ4 (Trustworthiness of Model Judgments).}
To explore this dimension, we assess whether:
\begin{inparaenum} [\bf(a)]
    \item LMs could reliably classify quality questions and
    \item SLMs serve as viable privacy-preserving evaluators.
\end{inparaenum}
We compared judgments for high quality (HQ) and skill match with a calibration view across strictness, sensitivity to pedagogical cues, and expert label agreement.

%%%%%%%%%%%%%%%%%%
\begin{table}[pt]
\centering
\tiny
\caption{Calibration view of SLMs, LLM, and Expert with HQ and Skill (S) (\%). Scores 2\% above expert benchmark are marked in green and 2\% below in red.}
\label{table:hq-sl-evaluations}
\vspace{-0.2cm}
\begin{tabular}{llcc|cc|cc|cc|ccllcc|cc|cc|cc|cc}
 &  & \multicolumn{2}{c|}{PS1} & \multicolumn{2}{c|}{PS2} & \multicolumn{2}{c|}{PS3} & \multicolumn{2}{c|}{PS4} & \multicolumn{2}{c}{PS5} &  &  & \multicolumn{2}{c|}{PS1} & \multicolumn{2}{c|}{PS2} & \multicolumn{2}{c|}{PS3} & \multicolumn{2}{c|}{PS4} & \multicolumn{2}{c}{PS5} \\ \cline{3-12} \cline{15-24} 
\begin{tabular}[c]{@{}l@{}}Eval\\ Model\end{tabular} & Model & \multicolumn{1}{l}{HQ} & \multicolumn{1}{l|}{S} & \multicolumn{1}{l}{HQ} & \multicolumn{1}{l|}{S} & \multicolumn{1}{l}{HQ} & \multicolumn{1}{l|}{S} & \multicolumn{1}{l}{HQ} & \multicolumn{1}{l|}{S} & \multicolumn{1}{l}{HQ} & \multicolumn{1}{l|}{S} & \begin{tabular}[c]{@{}l@{}}Eval\\ Model\end{tabular} & Model & \multicolumn{1}{l}{HQ} & \multicolumn{1}{l|}{S} & \multicolumn{1}{l}{HQ} & \multicolumn{1}{l|}{S} & \multicolumn{1}{l}{HQ} & \multicolumn{1}{l|}{S} & \multicolumn{1}{l}{HQ} & \multicolumn{1}{l|}{S} & \multicolumn{1}{l}{HQ} & \multicolumn{1}{l}{S} \\ \hline
Gr4 & G4 & \cellcolor[HTML]{FFCCC9}31 & \cellcolor[HTML]{FFCCC9}25 & \cellcolor[HTML]{FFCCC9}24 & \cellcolor[HTML]{FFCCC9}13 & \cellcolor[HTML]{FFCCC9}20 & \cellcolor[HTML]{FFCCC9}25 & \cellcolor[HTML]{FFCCC9}26 & \cellcolor[HTML]{FFCCC9}22 & \cellcolor[HTML]{FFCCC9}27 & \multicolumn{1}{c|}{\cellcolor[HTML]{FFCCC9}21} & P4 & G4 & \cellcolor[HTML]{FFCCC9}66 & \cellcolor[HTML]{FFCCC9}49 & \cellcolor[HTML]{FFCCC9}69 & \cellcolor[HTML]{FFCCC9}43 & \cellcolor[HTML]{FFCCC9}53 & \cellcolor[HTML]{FFCCC9}35 & \cellcolor[HTML]{FFCCC9}77 & \cellcolor[HTML]{FFCCC9}37 & \cellcolor[HTML]{FFCCC9}80 & \cellcolor[HTML]{FFCCC9}34 \\
 & G3.5 & \cellcolor[HTML]{FFCCC9}28 & \cellcolor[HTML]{FFCCC9}31 & \cellcolor[HTML]{FFCCC9}18 & \cellcolor[HTML]{FFCCC9}50 & \cellcolor[HTML]{FFCCC9}17 & \cellcolor[HTML]{FFCCC9}35 & \cellcolor[HTML]{FFCCC9}22 & \cellcolor[HTML]{FFCCC9}14 & \cellcolor[HTML]{FFCCC9}25 & \multicolumn{1}{c|}{\cellcolor[HTML]{FFCCC9}23} &  & G3.5 & \cellcolor[HTML]{FFCCC9}54 & \cellcolor[HTML]{FFCCC9}47 & \cellcolor[HTML]{FFCCC9}31 & \cellcolor[HTML]{FFCCC9}47 & \cellcolor[HTML]{FFCCC9}69 & \cellcolor[HTML]{FFCCC9}39 & \cellcolor[HTML]{FFCCC9}56 & \cellcolor[HTML]{FFCCC9}51 & \cellcolor[HTML]{FFCCC9}61 & \cellcolor[HTML]{FFCCC9}44 \\
 & P2 & \cellcolor[HTML]{FFCCC9}27 & \cellcolor[HTML]{FFCCC9}11 & \cellcolor[HTML]{9AFF99}98 & \cellcolor[HTML]{FFCCC9}31 & \cellcolor[HTML]{FFCCC9}32 & \cellcolor[HTML]{FFCCC9}30 & \cellcolor[HTML]{FFCCC9}35 & \cellcolor[HTML]{FFCCC9}31 & \cellcolor[HTML]{FFCCC9}24 & \multicolumn{1}{c|}{\cellcolor[HTML]{FFCCC9}21} &  & P2 & \cellcolor[HTML]{FFCCC9}54 & \cellcolor[HTML]{FFCCC9}42 & \cellcolor[HTML]{FFCCC9}45 & \cellcolor[HTML]{FFCCC9}37 & \cellcolor[HTML]{FFCCC9}71 & \cellcolor[HTML]{FFCCC9}40 & \cellcolor[HTML]{FFCCC9}63 & \cellcolor[HTML]{FFCCC9}39 & \cellcolor[HTML]{FFCCC9}57 & \cellcolor[HTML]{FFCCC9}31 \\
 & L2 & \cellcolor[HTML]{FFCCC9}33 & \cellcolor[HTML]{FFCCC9}24 & \cellcolor[HTML]{FFCCC9}21 & \cellcolor[HTML]{FFCCC9}33 & \cellcolor[HTML]{FFCCC9}25 & \cellcolor[HTML]{FFCCC9}20 & \cellcolor[HTML]{FFCCC9}26 & \cellcolor[HTML]{FFCCC9}30 & \cellcolor[HTML]{FFCCC9}25 & \multicolumn{1}{c|}{\cellcolor[HTML]{FFCCC9}36} &  & L2 & \cellcolor[HTML]{FFCCC9}70 & \cellcolor[HTML]{FFCCC9}37 & \cellcolor[HTML]{FFCCC9}50 & \cellcolor[HTML]{FFCCC9}43 & \cellcolor[HTML]{FFCCC9}45 & \cellcolor[HTML]{FFCCC9}30 & \cellcolor[HTML]{FFCCC9}50 & \cellcolor[HTML]{FFCCC9}37 & \cellcolor[HTML]{FFCCC9}43 & \cellcolor[HTML]{9AFF99}48 \\
 & M & \cellcolor[HTML]{FFCCC9}35 & \cellcolor[HTML]{FFCCC9}28 & \cellcolor[HTML]{FFCCC9}29 & \cellcolor[HTML]{FFCCC9}30 & \cellcolor[HTML]{FFCCC9}22 & \cellcolor[HTML]{FFCCC9}9 & \cellcolor[HTML]{FFCCC9}17 & \cellcolor[HTML]{FFCCC9}29 & \cellcolor[HTML]{FFCCC9}17 & \multicolumn{1}{c|}{\cellcolor[HTML]{FFCCC9}29} &  & M & \cellcolor[HTML]{FFCCC9}59 & \cellcolor[HTML]{FFCCC9}37 & \cellcolor[HTML]{FFCCC9}56 & \cellcolor[HTML]{FFCCC9}44 & \cellcolor[HTML]{FFCCC9}37 & \cellcolor[HTML]{FFCCC9}26 & \cellcolor[HTML]{FFCCC9}32 & \cellcolor[HTML]{FFCCC9}21 & \cellcolor[HTML]{FFCCC9}46 & \cellcolor[HTML]{FFCCC9}43 \\ \hline
P4m & G4 & \cellcolor[HTML]{FFCCC9}2 & \cellcolor[HTML]{FFCCC9}50 & \cellcolor[HTML]{FFCCC9}1 & \cellcolor[HTML]{FFCCC9}0 & \cellcolor[HTML]{FFCCC9}0 & \cellcolor[HTML]{FFCCC9}0 & \cellcolor[HTML]{FFCCC9}0 & \cellcolor[HTML]{FFCCC9}0 & \cellcolor[HTML]{FFCCC9}0 & \multicolumn{1}{c|}{\cellcolor[HTML]{FFCCC9}0} & D1 & G4 & \cellcolor[HTML]{9AFF99}91 & \cellcolor[HTML]{FFCCC9}69 & \cellcolor[HTML]{9AFF99}97 & \cellcolor[HTML]{FFCCC9}65 & \cellcolor[HTML]{9AFF99}98 & \cellcolor[HTML]{FFCCC9}57 & \cellcolor[HTML]{9AFF99}99 & \cellcolor[HTML]{FFCCC9}65 & 96 & \cellcolor[HTML]{FFCCC9}53 \\
 & G3.5 & \cellcolor[HTML]{FFCCC9}3 & \cellcolor[HTML]{FFCCC9}67 & \cellcolor[HTML]{FFCCC9}0 & \cellcolor[HTML]{FFCCC9}0 & \cellcolor[HTML]{FFCCC9}0 & \cellcolor[HTML]{FFCCC9}0 & \cellcolor[HTML]{FFCCC9}0 & \cellcolor[HTML]{FFCCC9}0 & \cellcolor[HTML]{FFCCC9}0 & \multicolumn{1}{c|}{\cellcolor[HTML]{FFCCC9}0} &  & G3.5 & \cellcolor[HTML]{9AFF99}93 & 81 & \cellcolor[HTML]{FFCCC9}90 & \cellcolor[HTML]{FFCCC9}63 & \cellcolor[HTML]{9AFF99}97 & \cellcolor[HTML]{FFCCC9}40 & \cellcolor[HTML]{9AFF99}95 & \cellcolor[HTML]{FFCCC9}56 & 93 & \cellcolor[HTML]{FFCCC9}56 \\
 & P2 & \cellcolor[HTML]{FFCCC9}2 & \cellcolor[HTML]{FFCCC9}50 & \cellcolor[HTML]{FFCCC9}1 & \cellcolor[HTML]{FFCCC9}0 & \cellcolor[HTML]{FFCCC9}0 & \cellcolor[HTML]{FFCCC9}0 & \cellcolor[HTML]{FFCCC9}0 & \cellcolor[HTML]{FFCCC9}0 & \cellcolor[HTML]{FFCCC9}0 & \multicolumn{1}{c|}{\cellcolor[HTML]{FFCCC9}0} &  & P2 & \cellcolor[HTML]{9AFF99}92 & \cellcolor[HTML]{9AFF99}67 & \cellcolor[HTML]{9AFF99}94 & \cellcolor[HTML]{9AFF99}73 & \cellcolor[HTML]{9AFF99}94 & \cellcolor[HTML]{FFCCC9}65 & \cellcolor[HTML]{9AFF99}94 & 69 & \cellcolor[HTML]{9AFF99}98 & \cellcolor[HTML]{9AFF99}53 \\
 & L2 & \cellcolor[HTML]{FFCCC9}0 & \cellcolor[HTML]{FFCCC9}0 & \cellcolor[HTML]{FFCCC9}1 & \cellcolor[HTML]{9AFF99}100 & \cellcolor[HTML]{FFCCC9}1 & \cellcolor[HTML]{FFCCC9}0 & \cellcolor[HTML]{FFCCC9}0 & \cellcolor[HTML]{FFCCC9}0 & \cellcolor[HTML]{FFCCC9}0 & \multicolumn{1}{c|}{\cellcolor[HTML]{FFCCC9}0} &  & L2 & \cellcolor[HTML]{9AFF99}93 & \cellcolor[HTML]{9AFF99}68 & \cellcolor[HTML]{9AFF99}89 & \cellcolor[HTML]{FFCCC9}68 & \cellcolor[HTML]{9AFF99}95 & \cellcolor[HTML]{FFCCC9}51 & \cellcolor[HTML]{9AFF99}93 & \cellcolor[HTML]{FFCCC9}63 & \cellcolor[HTML]{9AFF99}83 & \cellcolor[HTML]{9AFF99}58 \\
 & M & \cellcolor[HTML]{FFCCC9}1 & \cellcolor[HTML]{9AFF99}100 & \cellcolor[HTML]{FFCCC9}1 & \cellcolor[HTML]{FFCCC9}0 & \cellcolor[HTML]{FFCCC9}0 & \cellcolor[HTML]{FFCCC9}0 & \cellcolor[HTML]{FFCCC9}0 & \cellcolor[HTML]{FFCCC9}0 & \cellcolor[HTML]{FFCCC9}0 & \multicolumn{1}{c|}{\cellcolor[HTML]{FFCCC9}0} &  & M & \cellcolor[HTML]{9AFF99}88 & \cellcolor[HTML]{FFCCC9}52 & \cellcolor[HTML]{9AFF99}84 & \cellcolor[HTML]{FFCCC9}57 & \cellcolor[HTML]{9AFF99}87 & \cellcolor[HTML]{FFCCC9}39 & \cellcolor[HTML]{9AFF99}88 & \cellcolor[HTML]{FFCCC9}54 & \cellcolor[HTML]{9AFF99}81 & \cellcolor[HTML]{FFCCC9}51 \\ \hline
L3.2 & G4 & \cellcolor[HTML]{9AFF99}82 & \cellcolor[HTML]{FFCCC9}23 & \cellcolor[HTML]{FFCCC9}75 & \cellcolor[HTML]{FFCCC9}21 & \cellcolor[HTML]{FFCCC9}65 & \cellcolor[HTML]{FFCCC9}12 & \cellcolor[HTML]{FFCCC9}70 & \cellcolor[HTML]{FFCCC9}17 & \cellcolor[HTML]{FFCCC9}74 & \multicolumn{1}{c|}{\cellcolor[HTML]{FFCCC9}9} & Ge3 & G4 & \cellcolor[HTML]{FFCCC9}15 & \cellcolor[HTML]{FFCCC9}20 & \cellcolor[HTML]{FFCCC9}21 & \cellcolor[HTML]{FFCCC9}19 & \cellcolor[HTML]{FFCCC9}24 & \cellcolor[HTML]{FFCCC9}4 & \cellcolor[HTML]{FFCCC9}23 & \cellcolor[HTML]{FFCCC9}26 & \cellcolor[HTML]{FFCCC9}17 & \cellcolor[HTML]{FFCCC9}18 \\
 & G3.5 & \cellcolor[HTML]{9AFF99}82 & \cellcolor[HTML]{FFCCC9}23 & \cellcolor[HTML]{FFCCC9}59 & \cellcolor[HTML]{FFCCC9}28 & \cellcolor[HTML]{FFCCC9}60 & \cellcolor[HTML]{FFCCC9}21 & \cellcolor[HTML]{FFCCC9}65 & \cellcolor[HTML]{FFCCC9}14 & \cellcolor[HTML]{FFCCC9}69 & \multicolumn{1}{c|}{\cellcolor[HTML]{FFCCC9}16} &  & G3.5 & \cellcolor[HTML]{FFCCC9}26 & \cellcolor[HTML]{FFCCC9}37 & \cellcolor[HTML]{FFCCC9}10 & \cellcolor[HTML]{FFCCC9}10 & \cellcolor[HTML]{FFCCC9}15 & \cellcolor[HTML]{FFCCC9}53 & \cellcolor[HTML]{FFCCC9}17 & \cellcolor[HTML]{FFCCC9}24 & \cellcolor[HTML]{FFCCC9}11 & \cellcolor[HTML]{FFCCC9}9 \\
 & P2 & \cellcolor[HTML]{9AFF99}82 & \cellcolor[HTML]{FFCCC9}24 & \cellcolor[HTML]{9AFF99}81 & \cellcolor[HTML]{FFCCC9}22 & \cellcolor[HTML]{9AFF99}82 & \cellcolor[HTML]{FFCCC9}21 & 77 & \cellcolor[HTML]{FFCCC9}19 & \cellcolor[HTML]{9AFF99}88 & \multicolumn{1}{c|}{\cellcolor[HTML]{FFCCC9}20} &  & P2 & \cellcolor[HTML]{FFCCC9}17 & \cellcolor[HTML]{FFCCC9}53 & \cellcolor[HTML]{FFCCC9}12 & \cellcolor[HTML]{FFCCC9}33 & \cellcolor[HTML]{FFCCC9}11 & \cellcolor[HTML]{FFCCC9}45 & \cellcolor[HTML]{FFCCC9}9 & \cellcolor[HTML]{FFCCC9}22 & \cellcolor[HTML]{FFCCC9}13 & \cellcolor[HTML]{FFCCC9}23 \\
 & L2 & 72 & \cellcolor[HTML]{FFCCC9}21 & \cellcolor[HTML]{FFCCC9}66 & \cellcolor[HTML]{FFCCC9}21 & \cellcolor[HTML]{FFCCC9}54 & \cellcolor[HTML]{FFCCC9}16 & \cellcolor[HTML]{FFCCC9}56 & \cellcolor[HTML]{FFCCC9}12 & \cellcolor[HTML]{FFCCC9}45 & \multicolumn{1}{c|}{\cellcolor[HTML]{FFCCC9}9} &  & L2 & \cellcolor[HTML]{FFCCC9}18 & \cellcolor[HTML]{FFCCC9}50 & \cellcolor[HTML]{FFCCC9}20 & \cellcolor[HTML]{FFCCC9}30 & \cellcolor[HTML]{FFCCC9}25 & \cellcolor[HTML]{FFCCC9}35 & \cellcolor[HTML]{FFCCC9}19 & \cellcolor[HTML]{FFCCC9}26 & \cellcolor[HTML]{FFCCC9}20 & \cellcolor[HTML]{FFCCC9}20 \\
 & M & 71 & \cellcolor[HTML]{FFCCC9}19 & 75 & \cellcolor[HTML]{FFCCC9}17 & \cellcolor[HTML]{FFCCC9}59 & \cellcolor[HTML]{FFCCC9}18 & \cellcolor[HTML]{FFCCC9}56 & \cellcolor[HTML]{FFCCC9}16 & \cellcolor[HTML]{FFCCC9}55 & \multicolumn{1}{c|}{\cellcolor[HTML]{FFCCC9}20} &  & M & \cellcolor[HTML]{FFCCC9}15 & \cellcolor[HTML]{FFCCC9}27 & \cellcolor[HTML]{FFCCC9}9 & \cellcolor[HTML]{FFCCC9}56 & \cellcolor[HTML]{FFCCC9}19 & \cellcolor[HTML]{FFCCC9}32 & \cellcolor[HTML]{FFCCC9}21 & \cellcolor[HTML]{FFCCC9}33 & \cellcolor[HTML]{FFCCC9}19 & \cellcolor[HTML]{FFCCC9}32 \\ \hline
MS & G4 & \cellcolor[HTML]{9AFF99}98 & \cellcolor[HTML]{FFCCC9}37 & \cellcolor[HTML]{9AFF99}99 & \cellcolor[HTML]{FFCCC9}32 & \cellcolor[HTML]{9AFF99}99 & \cellcolor[HTML]{FFCCC9}31 & \cellcolor[HTML]{9AFF99}100 & \cellcolor[HTML]{FFCCC9}32 & \cellcolor[HTML]{9AFF99}99 & \multicolumn{1}{c|}{\cellcolor[HTML]{FFCCC9}33} & GP & G4 & \cellcolor[HTML]{9AFF99}80 & \cellcolor[HTML]{FFCCC9}38 & \cellcolor[HTML]{FFCCC9}75 & \cellcolor[HTML]{FFCCC9}35 & \cellcolor[HTML]{FFCCC9}77 & \cellcolor[HTML]{FFCCC9}38 & \cellcolor[HTML]{FFCCC9}74 & \cellcolor[HTML]{FFCCC9}40 & \cellcolor[HTML]{FFCCC9}67 & \cellcolor[HTML]{FFCCC9}38 \\
 & G3.5 & \cellcolor[HTML]{9AFF99}96 & \cellcolor[HTML]{FFCCC9}37 & \cellcolor[HTML]{9AFF99}100 & \cellcolor[HTML]{FFCCC9}26 & \cellcolor[HTML]{9AFF99}93 & \cellcolor[HTML]{FFCCC9}33 & \cellcolor[HTML]{9AFF99}95 & \cellcolor[HTML]{FFCCC9}30 & 93 & \multicolumn{1}{c|}{\cellcolor[HTML]{FFCCC9}34} &  & G3.5 & \cellcolor[HTML]{9AFF99}82 & \cellcolor[HTML]{FFCCC9}48 & \cellcolor[HTML]{FFCCC9}68 & \cellcolor[HTML]{FFCCC9}35 & \cellcolor[HTML]{FFCCC9}75 & \cellcolor[HTML]{FFCCC9}28 & \cellcolor[HTML]{FFCCC9}67 & \cellcolor[HTML]{FFCCC9}34 & \cellcolor[HTML]{FFCCC9}52 & \cellcolor[HTML]{FFCCC9}30 \\
 & P2 & \cellcolor[HTML]{9AFF99}92 & \cellcolor[HTML]{FFCCC9}36 & \cellcolor[HTML]{9AFF99}91 & \cellcolor[HTML]{FFCCC9}32 & \cellcolor[HTML]{9AFF99}98 & \cellcolor[HTML]{FFCCC9}33 & \cellcolor[HTML]{9AFF99}93 & \cellcolor[HTML]{FFCCC9}35 & \cellcolor[HTML]{9AFF99}90 & \multicolumn{1}{c|}{\cellcolor[HTML]{FFCCC9}33} &  & P2 & \cellcolor[HTML]{9AFF99}70 & \cellcolor[HTML]{FFCCC9}39 & \cellcolor[HTML]{FFCCC9}66 & \cellcolor[HTML]{FFCCC9}49 & 78 & \cellcolor[HTML]{FFCCC9}40 & 77 & \cellcolor[HTML]{FFCCC9}44 & 70 & 37 \\
 & L2 & \cellcolor[HTML]{9AFF99}97 & \cellcolor[HTML]{FFCCC9}32 & \cellcolor[HTML]{9AFF99}92 & \cellcolor[HTML]{FFCCC9}30 & \cellcolor[HTML]{FFCCC9}71 & \cellcolor[HTML]{FFCCC9}25 & \cellcolor[HTML]{9AFF99}81 & \cellcolor[HTML]{FFCCC9}31 & \cellcolor[HTML]{9AFF99}60 & \multicolumn{1}{c|}{\cellcolor[HTML]{FFCCC9}31} &  & L2 & \cellcolor[HTML]{9AFF99}79 & \cellcolor[HTML]{FFCCC9}40 & \cellcolor[HTML]{FFCCC9}66 & \cellcolor[HTML]{FFCCC9}43 & \cellcolor[HTML]{9AFF99}80 & \cellcolor[HTML]{FFCCC9}40 & \cellcolor[HTML]{FFCCC9}63 & \cellcolor[HTML]{FFCCC9}39 & \cellcolor[HTML]{9AFF99}54 & \cellcolor[HTML]{FFCCC9}35 \\
 & M & \cellcolor[HTML]{9AFF99}86 & \cellcolor[HTML]{FFCCC9}30 & \cellcolor[HTML]{9AFF99}89 & \cellcolor[HTML]{FFCCC9}27 & \cellcolor[HTML]{FFCCC9}69 & \cellcolor[HTML]{FFCCC9}27 & \cellcolor[HTML]{9AFF99}76 & \cellcolor[HTML]{FFCCC9}35 & \cellcolor[HTML]{9AFF99}72 & \multicolumn{1}{c|}{\cellcolor[HTML]{FFCCC9}32} &  & M & \cellcolor[HTML]{9AFF99}82 & \cellcolor[HTML]{FFCCC9}54 & \cellcolor[HTML]{FFCCC9}62 & \cellcolor[HTML]{FFCCC9}44 & \cellcolor[HTML]{FFCCC9}71 & \cellcolor[HTML]{FFCCC9}19 & \cellcolor[HTML]{FFCCC9}59 & \cellcolor[HTML]{FFCCC9}40 & \cellcolor[HTML]{FFCCC9}56 & \cellcolor[HTML]{FFCCC9}32 \\ \hline
GO & G4 & \cellcolor[HTML]{9AFF99}91 & \cellcolor[HTML]{9AFF99}89 & 85 & \cellcolor[HTML]{FFCCC9}76 & \cellcolor[HTML]{FFCCC9}81 & \cellcolor[HTML]{9AFF99}71 & \cellcolor[HTML]{FFCCC9}85 & \cellcolor[HTML]{9AFF99}90 & \cellcolor[HTML]{FFCCC9}77 & \multicolumn{1}{c|}{\cellcolor[HTML]{9AFF99}72} & Expert & G4 & 75 & 74 & 87 & 82 & 91 & 66 & 96 & 76 & 95 & 57 \\
 & G3.5 & \cellcolor[HTML]{9AFF99}86 & \cellcolor[HTML]{9AFF99}93 & \cellcolor[HTML]{FFCCC9}78 & \cellcolor[HTML]{FFCCC9}75 & \cellcolor[HTML]{FFCCC9}65 & 64 & \cellcolor[HTML]{FFCCC9}78 & \cellcolor[HTML]{FFCCC9}66 & \cellcolor[HTML]{FFCCC9}74 & \multicolumn{1}{c|}{\cellcolor[HTML]{9AFF99}75} &  & G3.5 & 70 & 82 & 94 & 90 & 89 & 65 & 87 & 73 & 91 & 59 \\
 & P2 & \cellcolor[HTML]{9AFF99}82 & \cellcolor[HTML]{9AFF99}71 & 75 & \cellcolor[HTML]{9AFF99}79 & \cellcolor[HTML]{9AFF99}88 & \cellcolor[HTML]{9AFF99}78 & \cellcolor[HTML]{9AFF99}84 & \cellcolor[HTML]{9AFF99}77 & \cellcolor[HTML]{9AFF99}78 & \multicolumn{1}{c|}{\cellcolor[HTML]{9AFF99}54} &  & P2 & 62 & 56 & 74 & 65 & 76 & 71 & 75 & 70 & 72 & 37 \\
 & L2 & \cellcolor[HTML]{9AFF99}88 & \cellcolor[HTML]{9AFF99}66 & 75 & \cellcolor[HTML]{9AFF99}80 & \cellcolor[HTML]{FFCCC9}67 & \cellcolor[HTML]{9AFF99}62 & \cellcolor[HTML]{FFCCC9}67 & 75 & \cellcolor[HTML]{9AFF99}56 & \multicolumn{1}{c|}{\cellcolor[HTML]{9AFF99}84} &  & L2 & 74 & 57 & 75 & 71 & 77 & 58 & 77 & 73 & 49 & 40 \\
 & M & \cellcolor[HTML]{9AFF99}81 & \cellcolor[HTML]{9AFF99}61 & \cellcolor[HTML]{FFCCC9}71 & \cellcolor[HTML]{FFCCC9}63 & \cellcolor[HTML]{FFCCC9}72 & \cellcolor[HTML]{FFCCC9}38 & \cellcolor[HTML]{FFCCC9}59 & \cellcolor[HTML]{9AFF99}77 & \cellcolor[HTML]{FFCCC9}62 & \multicolumn{1}{c|}{\cellcolor[HTML]{9AFF99}70} &  & M & 71 & 57 & 75 & 69 & 78 & 45 & 72 & 71 & 67 & 59
\end{tabular}
\vspace{-0.7cm}
\end{table}
%%%%%%%%%%%%%%%%%%

\noindent{\textit{Descriptive Calibration against Expert Judgments.}} We compared SLMs' judgments against the LLM and expert baseline (Table~\ref{table:hq-sl-evaluations}). Human experts consistently endorsed a heavy share \textit{($\approx$70--90\%)} as HQ and just over half \textit{$(51\%)$} among these as Skill-Matched. Comparatively, the LLM evaluator (GP) labeled a smaller proportion $(HQ\approx70\%; Skill\approx30\%)$. This indicated a conservative behavior overall and less attuned to Bloom-level distinctions, also hinting expert-level judgment characteristics with Palm (its sister model). On the other hand, SLMs showed substantial divergence, with some being too lenient and producing high HQ rates \textit{(shaded in green)} and others too conservative \textit{(shaded in red)}. 

%%%%%%%%%%%%%%%%%%%%%
\begin{table}[pt]
\tiny
\centering
\caption{Evaluation with Gold standard where \emph{A} - Accuracy, \emph{P} - Precision, \emph{R} - Recall, \emph{F} - F1 Score (in \%)}
\label{table:slm-gold-evaluation}
\vspace{-0.3cm}
\setlength{\tabcolsep}{3pt}
\begin{tabular}{ll|cccc|cccc|cccc|cccc|cccc}
 &  & \multicolumn{4}{c|}{PS1} & \multicolumn{4}{c|}{PS2} & \multicolumn{4}{c|}{PS3} & \multicolumn{4}{c|}{PS4} & \multicolumn{4}{c}{PS5} \\ \cline{3-22} 
\begin{tabular}[c]{@{}l@{}}Eval\\ Model\end{tabular} & Model & A & P & R & F & A & P & R & F & A & P & R & F & A & P & R & F & A & P & R & F \\ \hline
Gr4 & G4 & 59 & 97 & 43 & 60 & 36 & 96 & 26 & 41 & 30 & 95 & 21 & 35 & 28 & 96 & 27 & 42 & 30 & 96 & 28 & 43 \\
 & G3.5 & 51 & 83 & 35 & 49 & 26 & 94 & 19 & 31 & 27 & 100 & 19 & 31 & 31 & 86 & 22 & 35 & 30 & 92 & 26 & 40 \\
 & P2 & 52 & 75 & 33 & 46 & 44 & 80 & 32 & 46 & 42 & 82 & 34 & 48 & 38 & 58 & 30 & 40 & 50 & 88 & 30 & 45 \\
 & L2 & 67 & 0 & 0 & 0 & 39 & 76 & 22 & 34 & 57 & 84 & 34 & 49 & 49 & 67 & 30 & 41 & 69 & 72 & 42 & 53 \\
 & M & 65 & 0 & 0 & 0 & 51 & 83 & 36 & 50 & 54 & 77 & 29 & 42 & 53 & 76 & 23 & 35 & 45 & 65 & 18 & 28 \\ \hline
P4m & G4 & 31 & 100 & 3 & 5 & 16 & 100 & 1 & 2 & 13 & 0 & 0 & 0 & 4 & 0 & 0 & 0 & 5 & 0 & 0 & 0 \\
 & G3.5 & 33 & 67 & 3 & 6 & 11 & 0 & 0 & 0 & 11 & 0 & 0 & 0 & 16 & 0 & 0 & 0 & 9 & 0 & 0 & 0 \\
 & P2 & 40 & 100 & 3 & 6 & 27 & 100 & 1 & 3 & 22 & 0 & 0 & 0 & 32 & 0 & 0 & 0 & 32 & 0 & 0 & 0 \\
 & L2 & 100 & 0 & 0 & 0 & 27 & 0 & 0 & 0 & 41 & 100 & 2 & 3 & 40 & 0 & 0 & 0 & 58 & 0 & 0 & 0 \\
 & M & 99 & 0 & 0 & 0 & 32 & 100 & 1 & 3 & 42 & 0 & 0 & 0 & 44 & 0 & 0 & 0 & 40 & 0 & 0 & 0 \\ \hline
L3.2 & G4 & 71 & 75 & 88 & 81 & 74 & 89 & 78 & 83 & 64 & 89 & 66 & 76 & 72 & 99 & 71 & 83 & 73 & 96 & 74 & 84 \\
 & G3.5 & 60 & 67 & 81 & 73 & 58 & 90 & 59 & 72 & 63 & 93 & 63 & 75 & 67 & 89 & 69 & 78 & 70 & 94 & 71 & 81 \\
 & P2 & 66 & 67 & 89 & 76 & 71 & 77 & 85 & 81 & 71 & 80 & 84 & 82 & 61 & 68 & 78 & 73 & 66 & 69 & 90 & 78 \\
 & L2 & 28 & 0 & 0 & 0 & 67 & 79 & 73 & 76 & 55 & 64 & 57 & 60 & 57 & 65 & 61 & 63 & 75 & 70 & 74 & 72 \\
 & M & 29 & 0 & 0 & 0 & 71 & 76 & 83 & 79 & 68 & 72 & 73 & 72 & 53 & 58 & 58 & 58 & 64 & 71 & 66 & 68 \\ \hline
MS & G4 & 71 & 71 & 99 & 83 & 86 & 86 & 100 & 93 & 88 & 88 & 100 & 94 & 96 & 96 & 100 & 98 & 94 & 95 & 99 & 97 \\
 & G3.5 & 66 & 67 & 96 & 79 & 89 & 89 & 100 & 94 & 88 & 92 & 96 & 94 & 85 & 87 & 98 & 92 & 88 & 93 & 95 & 94 \\
 & P2 & 68 & 66 & 98 & 79 & 75 & 76 & 95 & 85 & 80 & 80 & 100 & 89 & 69 & 69 & 96 & 80 & 68 & 70 & 93 & 80 \\
 & L2 & 3 & 0 & 0 & 0 & 72 & 73 & 95 & 83 & 74 & 74 & 87 & 80 & 71 & 69 & 93 & 79 & 76 & 66 & 93 & 77 \\
 & M & 14 & 0 & 0 & 0 & 74 & 74 & 96 & 83 & 75 & 74 & 88 & 81 & 62 & 62 & 84 & 71 & 78 & 77 & 92 & 84 \\ \hline
GO & G4 & 74 & 74 & 96 & 84 & 82 & 90 & 90 & 90 & 76 & 89 & 83 & 86 & 83 & 97 & 86 & 91 & 75 & 95 & 77 & 85 \\
 & G3.5 & 66 & 69 & 88 & 78 & 75 & 91 & 80 & 85 & 68 & 94 & 68 & 79 & 69 & 84 & 78 & 81 & 73 & 93 & 75 & 83 \\
 & P2 & 70 & 69 & 92 & 79 & 79 & 86 & 87 & 86 & 76 & 81 & 91 & 86 & 64 & 69 & 86 & 76 & 56 & 65 & 75 & 70 \\
 & L2 & 12 & 0 & 0 & 0 & 64 & 74 & 77 & 75 & 60 & 65 & 72 & 68 & 58 & 63 & 70 & 67 & 55 & 47 & 63 & 54 \\
 & M & 19 & 0 & 0 & 0 & 76 & 82 & 84 & 83 & 57 & 60 & 75 & 67 & 58 & 62 & 65 & 63 & 65 & 70 & 72 & 71 \\ \hline
P4 & G4 & 77 & 87 & 81 & 83 & 72 & 91 & 74 & 82 & 60 & 94 & 57 & 71 & 79 & 99 & 80 & 88 & 81 & 98 & 82 & 89 \\
 & G3.5 & 65 & 80 & 64 & 71 & 40 & 97 & 34 & 50 & 74 & 96 & 74 & 83 & 64 & 93 & 62 & 74 & 62 & 94 & 62 & 75 \\
 & P2 & 76 & 85 & 75 & 80 & 64 & 91 & 56 & 69 & 75 & 88 & 79 & 83 & 64 & 75 & 70 & 72 & 70 & 83 & 70 & 76 \\
 & L2 & 30 & 0 & 0 & 0 & 55 & 76 & 53 & 63 & 70 & 83 & 62 & 71 & 63 & 73 & 61 & 66 & 77 & 73 & 74 & 74 \\
 & M & 41 & 0 & 0 & 0 & 72 & 86 & 70 & 77 & 66 & 82 & 53 & 64 & 61 & 76 & 44 & 56 & 73 & 85 & 66 & 74 \\ \hline
D1 & G4 & 75 & 75 & 97 & 85 & 84 & 86 & 98 & 91 & 85 & 87 & 98 & 92 & 95 & 96 & 99 & 97 & 93 & 96 & 97 & 96 \\
 & G3.5 & 69 & 69 & 96 & 80 & 89 & 93 & 95 & 94 & 86 & 89 & 97 & 93 & 85 & 87 & 98 & 92 & 86 & 92 & 94 & 93 \\
 & P2 & 68 & 66 & 98 & 79 & 72 & 74 & 95 & 83 & 78 & 80 & 96 & 88 & 66 & 68 & 94 & 79 & 66 & 67 & 97 & 79 \\
 & L2 & 7 & 0 & 0 & 0 & 73 & 75 & 93 & 83 & 63 & 62 & 98 & 76 & 65 & 63 & 98 & 77 & 55 & 48 & 95 & 64 \\
 & M & 12 & 0 & 0 & 0 & 75 & 76 & 93 & 83 & 61 & 61 & 92 & 73 & 64 & 61 & 96 & 75 & 73 & 70 & 95 & 81 \\ \hline
Ge3 & G4 & 30 & 53 & 11 & 18 & 29 & 86 & 21 & 33 & 26 & 79 & 21 & 34 & 25 & 96 & 22 & 36 & 22 & 100 & 18 & 30 \\
 & G3.5 & 41 & 67 & 26 & 38 & 19 & 90 & 10 & 18 & 20 & 80 & 13 & 23 & 25 & 76 & 15 & 25 & 18 & 91 & 11 & 19 \\
 & P2 & 24 & 6 & 2 & 3 & 34 & 83 & 13 & 23 & 26 & 73 & 10 & 18 & 33 & 56 & 7 & 13 & 35 & 62 & 12 & 20 \\
 & L2 & 82 & 0 & 0 & 0 & 30 & 55 & 15 & 24 & 44 & 58 & 25 & 34 & 41 & 53 & 16 & 25 & 54 & 40 & 19 & 25 \\
 & M & 85 & 0 & 0 & 0 & 34 & 67 & 9 & 15 & 39 & 42 & 14 & 21 & 41 & 43 & 16 & 23 & 39 & 47 & 15 & 23
\end{tabular}
\vspace{-0.7cm}
\end{table}
%%%%%%%%%%%%%%%%%%%%%%

\noindent{\textit{Agreement with Expert Labels.}} We then computed agreement-based evaluation with expert labels (Table~\ref{table:slm-gold-evaluation}). Results revealed clear differences, with some consistently being the strongest evaluators, particularly with frontier model questions, and few performing poorly with low F1 scores (<50\%), indicating degenerate behavior of almost no HQ items. ANOVA confirmed these differences with post-hoc analysis, grouping some models (D1, MS, GO, P4, and L3.2) above others.

\noindent\textit{Model Bias in Evaluation.} Performance was not uniform even among the stronger SLMs. Some F1 scores lay between \textit{80–97\%} but were lower and more variable for others. Similarly, patterns of scoring frontier models more consistently appeared. These inconsistencies revealed that evaluator reliability depended not only on the evaluator but also on the model being judged, with favorable behavior emerging. This underscores caution for deploying these evaluators at scale.

These results show that neither SLMs nor LLMs replicate the expert's balance of selectivity and skill awareness. Even quantitative analysis showcased substantial variance with:
\begin{inparaenum} [\it a.]
    \item some over-accepting questions \textit{(risking low-quality items)} and
    \item some under-accepting them \textit{(risking removal of potentially high-quality items)}.
\end{inparaenum}
This variability indicated model-induced bias with no generalizable results.
%%%%%%%%%%%%%%%%%%%%%%%%%%%%%%%%%%%%%%%%%%%%%%%%%%%%%%%%%%%%%%%%%
\section{Discussion}
\label{sec:disc}
%%%%%
\noindent{\textit{LM Generation Capabilities under Pedagogical Constraints.}} This study's goal is not to claim pedagogical correctness or learning impact, but to provide a rigorous, deployment-aware evaluation framework for assessing how generative models behave under pedagogical constraints in realistic instructor-facing workflows. Consistent with LLM-based educational content generation works~\cite{scaria2024automated,wang2022towards,raz2025automated}, we primarily examined LMs' use for AEQG under explicit pedagogical constraints (Bloom's taxonomy), where large models generally produced high-fidelity questions aligned with intended Bloom levels and prompt specifications when instructional cues were sufficiently explicit. Yet, a notable finding was that SLMs, despite substantially smaller parameter counts and resource footprints, achieved comparable performance. This aligns with evidence that model size alone is not a reliable predictor of effectiveness, especially for constrained tasks, which was an observed limitation of LLMs~\cite{tam2024let,bulathwela2023scalable}. SLMs, however, exhibited lower variance across PS and more consistent compliance with Bloom's constraints, suggesting that smaller models may encode instructional constraints in a more conservative and predictable manner. This stability is especially relevant in instructional contexts where prompts are authored iteratively by educators rather than optimized by expert prompt engineers. 

\noindent{\textit{Assessment Validity Considerations.}} From an assessment theory perspective, the observed trade-offs can be interpreted through an argument-based validity lens, where evidence is accumulated to justify the intended interpretation and use of assessment items. In this framing, Bloom alignment and topical adherence provide partial evidence for \emph{content relevance} and \emph{construct representation}, while readability and grade-level proxies relate to \emph{accessibility} and potential construct-irrelevant variance. Our results, therefore, support a cautious validity argument: LMs can assist with item drafting under explicit constraints, but the remaining variability and occasional drift imply that human review remains necessary before items are used in decisions.

\noindent{\textit{Trust and Limits of Model-based Evaluation.}} While generation performance was strong across both model families, evaluation quality using LMs revealed greater fragility. Though some LMs achieved moderate-to-high agreement with experts, their behavior varied with distinct patterns of inconsistent skill recognition and over- or under-acceptance, consistent with model-as-a-judge pitfalls of prompt sensitivity, positional bias, and calibration errors~\cite{zheng2023judging,fawzi2024towards}. In high-stakes educational contexts, even small inconsistencies could lead to misleading assessments or unfair filtering. This work extends this to pedagogically structured educational tasks, demonstrating that even well-defined evaluation targets can diverge automated judgments meaningfully from expert reasoning. Additionally, evaluator-specific biases, in which some exhibited preferential agreement with particular same-family generators, raise concerns about fairness and comparability and against unmoderated use of AI-based evaluators in educational contexts~\cite{umokegovernance}. On the other hand, reliance on purely subjective rubric-based evaluation itself may be insufficient, as pedagogical quality often involves nuanced tradeoffs that models often fail to capture through rigid categorical criteria alone~\cite{elkins2023useful}. Together, these warrant further research on methodological rigor, especially in educational AI, where the same human-tested techniques may not apply, requiring a combination of reproducible metrics with expert judgment.

\noindent{\textit{Implications for Instructional Use and Deployment.}} From a practical perspective, several SLMs' more frequent stability has important implications for instructional deployment, with their highly compliant performances indicating suitability for real-world educational settings where prompts are often brief, under-specified, or instructor-generated. It also aligns with recent calls for scalable, lightweight educational AI systems that can operate under privacy, governance, and cost constraints~\cite{bulathwela2023scalable}. At the same time, trends in readability and grade level complexity suggest that high-requirement prompts naturally induce more linguistically complex outputs, especially at higher cognition-demanding Bloom's levels. While this behavior is pedagogically interpretable and consistent with cognitive demand and language complexity findings~\cite{armstrong2010bloom,raz2025automated}, it reinforces the need for human oversight to ensure accessibility for the target learner. Any automated system employing solely cognitive alignment risks introducing unintended barriers to comprehension, especially in diverse or mixed-ability classrooms.

\noindent{\textit{Constructive Alignment and Instructional Intent.}} These findings also align with the principle of constructive alignment that assessment prompts should elicit evidence aligned with intended learning outcomes and that the resulting items should remain faithful to the targeted cognitive demand. Our Bloom-conditioned generation setup operationalizes this alignment at the prompt level, while the topical and lexical compliance checks provide scalable signals of whether generated items preserve instructional intent across levels. The fact that stronger prompting generally improves alignment further reinforces that item quality is jointly produced by model behavior and the instructional specification supplied by educators.

\noindent{\textit{Potential as Bounded Pedagogical Assistants.}} At the current state, the most viable use suggests that LMs should be used as \emph{bounded assistants} rather than autonomous assessment designers and complete replacements for human expertise. Both model families demonstrated meaningful capacity to support AEQG, but neither consistently satisfied all requirements without oversight. SLMs, however, offer an interesting avenue for these purposes, especially due to their lower computational demands and feasibility for on-device and small school-based deployments, offering a sustainable pathway, especially as these environments usually have significant privacy concerns, equity of access, and cost effectiveness. Yet, variability in generation results and evaluation reliability, coupled with model-specific biases, underscores the necessity of working towards a semi-automated pipeline in which a trustworthy moderator is always in the loop, popularly known as HIL frameworks~\cite{memarian2024human}, for responsible adoption, particularly in curriculum-aligned and high-stakes assessment contexts. In this framing, they serve as tools for accelerating ideation, scaffolding question construction, and screening candidate items prior to human confirmation, while instructors and subject-matter experts retain curatorial authority. Such workflows balance efficiency with accountability and align with broader recommendations for responsible adoption in education~\cite{kasneci2023chatgpt}. Rather than replacing instructional expertise, these models can augment educator processes when embedded within transparent, moderated systems that preserve pedagogical intent and fairness.

\noindent{\textbf{Limitations}.} Although this study provides many insightful findings, there exist several limitations that can temper its generalizability. First, it was confined to a specific subject domain (data science), which constrains some generalization. Our evaluation deliberately supplied only minimal topic-level prompts and does not include optimization mechanisms such as fine-tuning. As a result, findings reflect each LM's baseline pre-training knowledge rather than leveraging content from curricular materials. The validation target group was also relatively homogeneous, which may limit the diversity of pedagogical perspectives. While expert judgment provides a reference point, it is not free from subjectivity. Prompting strategies, although varied significantly in our study, cannot exhaustively capture the diverse ways instructors might interact with the tools in authentic classroom workflows. Additionally, assessing Bloom's compliance with a verb check, although commonly used~\cite{adams2015bloom}, is a surface-level proxy, focusing only on specific term usage, and may overlook deeper alignment between the question, targeted skills, and pedagogical level. Additionally, ethical considerations, particularly those related to bias and fairness, are important when deploying AEQG pipelines. However, these issues fall outside the scope of the present study. Finally, some metrics (e.g., Grade Level and Reading Ease) are mathematically related and should be interpreted descriptively rather than as independent variables, as they reflect surface text complexity and not instructional clarity.

%%%%%%%%%%%%%%%%%%%%%%%%%%%%%%%%%%%%%%%%%%%%%%%%%%%%%%%%%%%%%%%%%
\section{Conclusion}
\label{sec:conc}
%%%%%
We demonstrate that while frontier LLMs project to offer high generational capacity, SLMs represent a promising alternative for locally deployed, privacy-preserving educational tools, especially with automated generation via reproducible multi-dimensional NLP checks and self-assessment tasks, which shows promise for reducing instructor workload. However, the trustworthiness of AI as evaluators remains unsettled, which further underscores the necessity of human oversight. With grounded deployment alongside an HIL framework and further research on fairness, reliability, and classroom practices, we can chart the path of responsible GenAI integration into education in a way that safeguards both efficiency and pedagogical integrity.

\noindent{\textit{Future Work.}}
Future research describes a roadmap towards responsible integration of AI in education. This includes expanding subject domains and grade-level coverage, diversifying multi-expert evaluation, conducting thorough pedagogical validity checks, and exploring fine-tuned and reinforcement-enabled generators alongside evaluators that are instruction-tuned for educational settings. Addressing evaluator biases, both by fairer model design and multiple evaluator combinations, can better triangulate judgments and robustness. Classroom-based deployment studies are needed to examine many intricate details about how instructors and learners actually interact with GenAI systems in practice and what unpredictable results these interactions may trigger. This analysis could provide insights into both efficacy and unintended consequences, implications for development of HIL-enabled decision-making pipelines, and hybrid evaluation of these systems~\cite{ilkou2024hybrid}. 

%%%% Acks
\begin{credits}
\subsubsection{\ackname} Chris Davis Jaldi and Cogan Shimizu acknowledge support from National Science Foundation under awards \#2333532 ``Proto-OKN Theme 3: An Education Gateway for the Proto-OKN (EduGate)'' and \#2537485 ``Prototyping a new Knowledge Resource for modern AI (Proto-KAI).'' Anmol Saini and Cogan Shimizu acknowledge support from DAGSI RX24-22. Noah Schroeder and Shan Zhang acknowledge support from the National Science Foundation and Institute for Education Sciences under Grant DRL-2229612. Any opinions, findings, and conclusions or recommendations expressed in this material are those of the author(s) and do not necessarily reflect the views of National Science Foundation or the U.S. Department of Education.
\end{credits}
%%%%%%%%%%%%%%%%%%%%%%%%%%%%%%%%%%%%%%%%%%%%%%%%%%%%%%%%%%%%%%%%%
\bibliographystyle{splncs04}
\bibliography{refs}

\begin{thebibliography}{10}
\providecommand{\url}[1]{\texttt{#1}}
\providecommand{\urlprefix}{URL }
\providecommand{\doi}[1]{https://doi.org/#1}

\bibitem{adams2015bloom}
Adams, N.E.: Bloom’s taxonomy of cognitive learning objectives. Journal of the Medical Library Association: JMLA  \textbf{103}(3), ~152 (2015)

\bibitem{allison2025generative}
Allison, J., Hwang, G.J., Mayer, R.E., Pellas, N., et~al.: From generative ai to extended reality: Multidisciplinary perspectives on the challenges, opportunities, and future of educational computing (2025)

\bibitem{armstrong2010bloom}
Armstrong, P.: Bloom’s taxonomy. Vanderbilt University Center for Teaching  \textbf{12}(05), ~2023 (2010)

\bibitem{site:git-repo}
{Blooms Taxonomy LM Analysis}. \url{https://github.com/kastle-lab/bloom-taxonomy-lm-analysis}, accessed: 2026-01-21

\bibitem{borchers2025can}
Borchers, C., Shou, T.: Can large language models match tutoring system adaptivity? a benchmarking study. In: AIED (2025)

\bibitem{bulathwela2023scalable}
Bulathwela, S., Muse, H., Yilmaz, E.: Scalable educational question generation with pre-trained language models. In: AIED (2023)

\bibitem{elkins2023useful}
Elkins, S., Kochmar, E., Serban, I., Cheung, J.C.: How useful are educational questions generated by large language models? In: International Conference on Artificial Intelligence in Education. pp. 536--542 (2023)

\bibitem{fawzi2024towards}
Fawzi, F., Balan, S., et~al.: Towards human-like educational question generation with small language models. In: AIED (2024)

\bibitem{flesch1950measuring}
Flesch, R.: Measuring the level of abstraction. Journal of applied psychology  \textbf{34}(6), ~384 (1950)

\bibitem{flesch1948new}
Flesch, R.: A new readability yardstick. Journal of applied psychology  \textbf{32}(3), ~221 (1948)

\bibitem{hwang2024towards}
Hwang, K., Wang, K., et~al.: Towards automated multiple choice question generation and evaluation: aligning with bloom’s taxonomy. In: AIED (2024)

\bibitem{ilkou2025dyslexia}
Ilkou, E., Alexiou, T., Antoniou, G., Viberg, O.: Dyslexia and ai: Do language models align with dyslexic style guide criteria? In: AIED (2025)

\bibitem{ilkou2024hybrid}
Ilkou, E., Linzbach, S., Wallat, J.: Hybrid evaluation of socratic questioning for teaching. In: Proceedings of ISWC'24 (2024)

\bibitem{jaldi2025impact}
Jaldi, C.D.: Impact of Graph Structures for RAG Outcomes in LLMs. Master's thesis, Wright State University (2025)

\bibitem{jaldi2025education}
Jaldi, C.D., Ilkou, E., Schroeder, N., Shimizu, C.: Education in the era of neurosymbolic ai. Journal of Web Semantics  \textbf{85},  100857 (2025)

\bibitem{kabir2023llm}
Kabir, M.R., Lin, F.: An llm-powered adaptive practicing system. In: LLM@ AIED. pp. 43--52 (2023)

\bibitem{kasneci2023chatgpt}
Kasneci, E., et~al.: Chatgpt for good? on opportunities and challenges of large language models for education. Learning and individual differences  (2023)

\bibitem{maity2024exploring}
Maity, S., Deroy, A., Sarkar, S.: Exploring the capabilities of prompted large language models in educational and assessment applications. arXiv preprint arXiv:2405.11579  (2024)

\bibitem{memarian2024human}
Memarian, B., Doleck, T.: Human-in-the-loop in artificial intelligence in education: A review and entity-relationship (er) analysis. Computers in Human Behavior: Artificial Humans  \textbf{2}(1),  100053 (2024)

\bibitem{meyer2024using}
Meyer, J., et~al.: Using llms to bring evidence-based feedback into the classroom: Ai-generated feedback increases secondary students’ text revision, motivation, and positive emotions. Computers and Education: Artificial Intelligence  (2024)

\bibitem{raz2025automated}
Raz, T., Luchini, S.A., et~al.: Automated scoring of question complexity with transformer language models. Thinking Skills and Creativity p. 102090 (2025)

\bibitem{reza2025small}
Reza, Z., Mazur, A., Dugdale, M., Ray-Chaudhuri, R.: Small models, big support: A local llm framework for teacher-centric content creation and assessment using rag and cag. arXiv preprint arXiv:2506.05925  (2025)

\bibitem{scaria2024automated}
Scaria, N., Dharani~Chenna, S., et~al.: Automated educational question generation at different bloom’s skill levels using large language models: Strategies and evaluation. In: International Conference on Artificial Intelligence in Education (2024)

\bibitem{schick2020s}
Schick, T., Sch{\"u}tze, H.: It's not just size that matters: Small language models are also few-shot learners. arXiv preprint arXiv:2009.07118  (2020)

\bibitem{stamper2024enhancing}
Stamper, J., Xiao, R., Hou, X.: Enhancing llm-based feedback: Insights from intelligent tutoring systems and the learning sciences. In: AIED (2024)

\bibitem{tam2024let}
Tam, Z.R., Wu, Lin, et~al.: Let me speak freely? a study on the impact of format restrictions on performance of large language models. arXiv preprint arXiv:2408.02442  (2024)

\bibitem{umokegovernance}
Umoke, C.C., Nwangbo, S.O., Onwe, O.A.: The governance of ai in education: Developing ethical policy frameworks for adaptive learning technologies

\bibitem{vanzo2024gpt}
Vanzo, A., Chowdhury, S.P., et~al.: Gpt-4 as a homework tutor can improve student engagement and learning outcomes. arXiv preprint arXiv:2409.15981  (2024)

\bibitem{vuruma2024cloud}
Vuruma, S.K.R., Margetts, A., Su, J., Ahmed, F., Srivastava, B.: From cloud to edge: Rethinking generative ai for low-resource design challenges. arXiv preprint arXiv:2402.12702  (2024)

\bibitem{wang2022towards}
Wang, Z., Valdez, J., et~al.: Towards human-like educational question generation with large language models. In: AIED (2022)

\bibitem{zhang2024transforming}
Zhang, H., Leong, W.: Transforming rural and underserved schools with ai-powered education solutions. ASM Science Journal  \textbf{19}, ~1895 (2024)

\bibitem{zhang2024semi}
Zhang, S., Palaguachi, C., Pitera, M., Jaldi, C.D., Schroeder, N.L., et~al.: Semi-automating the scoping review process: Is it worthwhile? a methodological evaluation. Educational Psychology Review  (2024)

\bibitem{zhang2019bertscore}
Zhang, T., Kishore, V., Wu, F., Weinberger, K.Q., Artzi, Y.: Bertscore: Evaluating text generation with bert. arXiv preprint arXiv:1904.09675  (2019)

\bibitem{zheng2023judging}
Zheng, L., et~al.: Judging llm-as-a-judge with mt-bench and chatbot arena. Advances in neural information processing systems  (2023)

\end{thebibliography}
%%%%%%%%%%%%%%%%%%%%%%%%%%%%%%%%%%%%%%%%%%%%%%%%%%%%%%%%%%%%%%%%%
\end{document}